\documentclass[10pt,twocolumn,letterpaper]{article}

\usepackage{cvpr}              

\usepackage{graphicx}
\usepackage{amsmath}
\usepackage{amssymb}
\usepackage{booktabs}
\usepackage{fancyvrb}
\usepackage{amsfonts}
\usepackage[thinc]{esdiff}
\usepackage{enumitem}
\usepackage[subtle]{savetrees}
\usepackage{siunitx}
\usepackage[accsupp]{axessibility}

\usepackage[pagebackref,breaklinks,colorlinks]{hyperref}

\usepackage[capitalize]{cleveref}
\crefname{section}{Sec.}{Secs.}
\Crefname{section}{Section}{Sections}
\Crefname{table}{Table}{Tables}
\crefname{table}{Tab.}{Tabs.}

\def\cvprPaperID{9} 
\def\confName{CVPR}
\def\confYear{2022}

\begin{document}


\title{MonoTrack: Shuttle trajectory reconstruction from monocular badminton video}

\author{Paul Liu\\
Stanford University\\
Stanford, CA\\
{\tt\small paul.liu@stanford.edu}
\and
Jui-Hsien Wang\\
Adobe Research\\
Seattle, WA\\
{\tt\small juiwang@adobe.com}
}
\maketitle

\begin{abstract}

Trajectory estimation is a fundamental component of racket sport analytics, as the trajectory contains information not only about the winning and losing of each point, but also how it was won or lost. In sports such as badminton, players benefit from knowing the full 3D trajectory, as the height of shuttlecock or ball provides valuable tactical information. Unfortunately, 3D reconstruction is a notoriously hard problem, and standard trajectory estimators can only track 2D pixel coordinates. In this work, we present the first complete end-to-end system for the extraction and segmentation of 3D shuttle trajectories from monocular badminton videos. Our system integrates badminton domain knowledge such as court dimension, shot placement, physical laws of motion, along with vision-based features such as player poses and shuttle tracking. We find that significant engineering efforts and model improvements are needed to make the overall system robust, and as a by-product of our work, improve state-of-the-art results on court recognition, 2D trajectory estimation, and hit recognition.

\end{abstract}

\section{Introduction}
\label{sec:intro}


Badminton is the world's second largest sport by participation~\cite{badminton_participation}. However, compared to more celebrated sports such as soccer and tennis, badminton has yet to enjoy the recent deep learning advances in computer vision. This is not due to the lack of need. Key analytic metrics such as trajectory information are so important that teams and athletes painstakingly annotate tournament matches and training videos manually. The result justifies the cause, as Carolina Marin, an Olympic gold medallist in badminton and three-time world champion, was rumored to hire a team of 6 annotators for labeling all of her matches. In this work, we aim to help athletes, coaches, and hobbyists alike in reducing the manual labor required to label badminton videos. 

\begin{figure}
    \centering
    \includegraphics[height=0.1\textwidth, width=0.32\columnwidth]{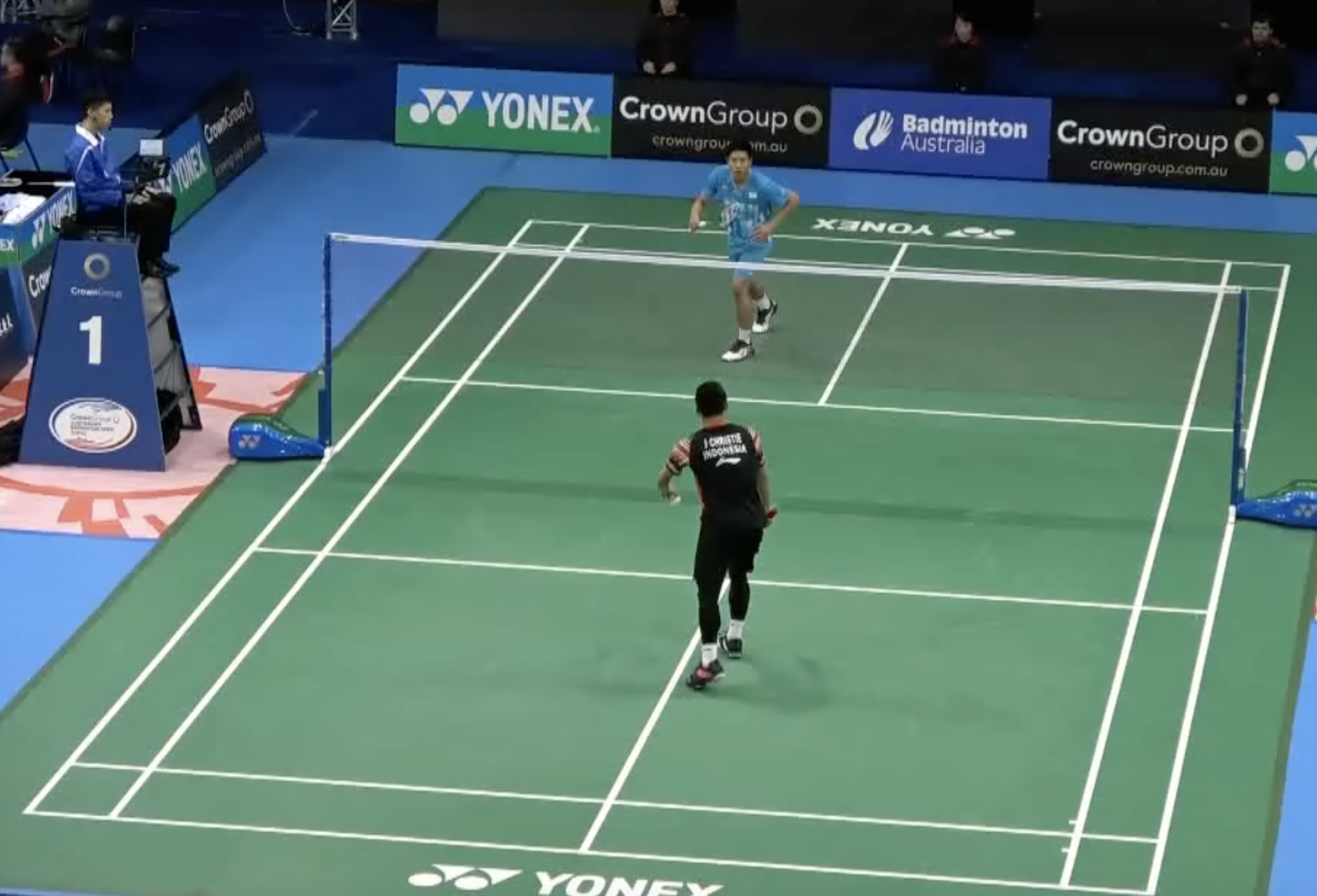}\linebreak[0]
    \includegraphics[height=0.1\textwidth, width=0.32\columnwidth]{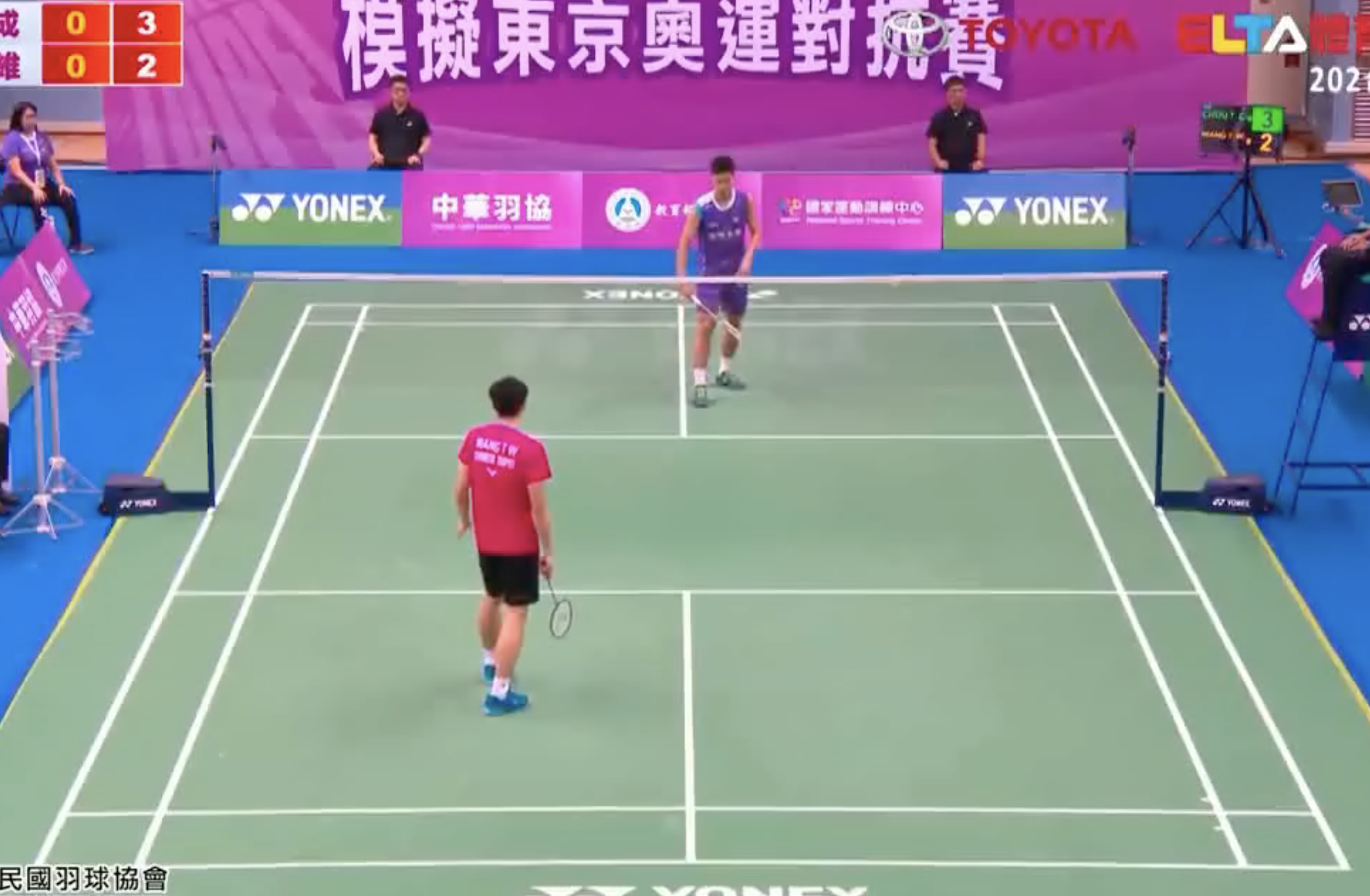}\linebreak[0]
    \includegraphics[height=0.1\textwidth, width=0.32\columnwidth]{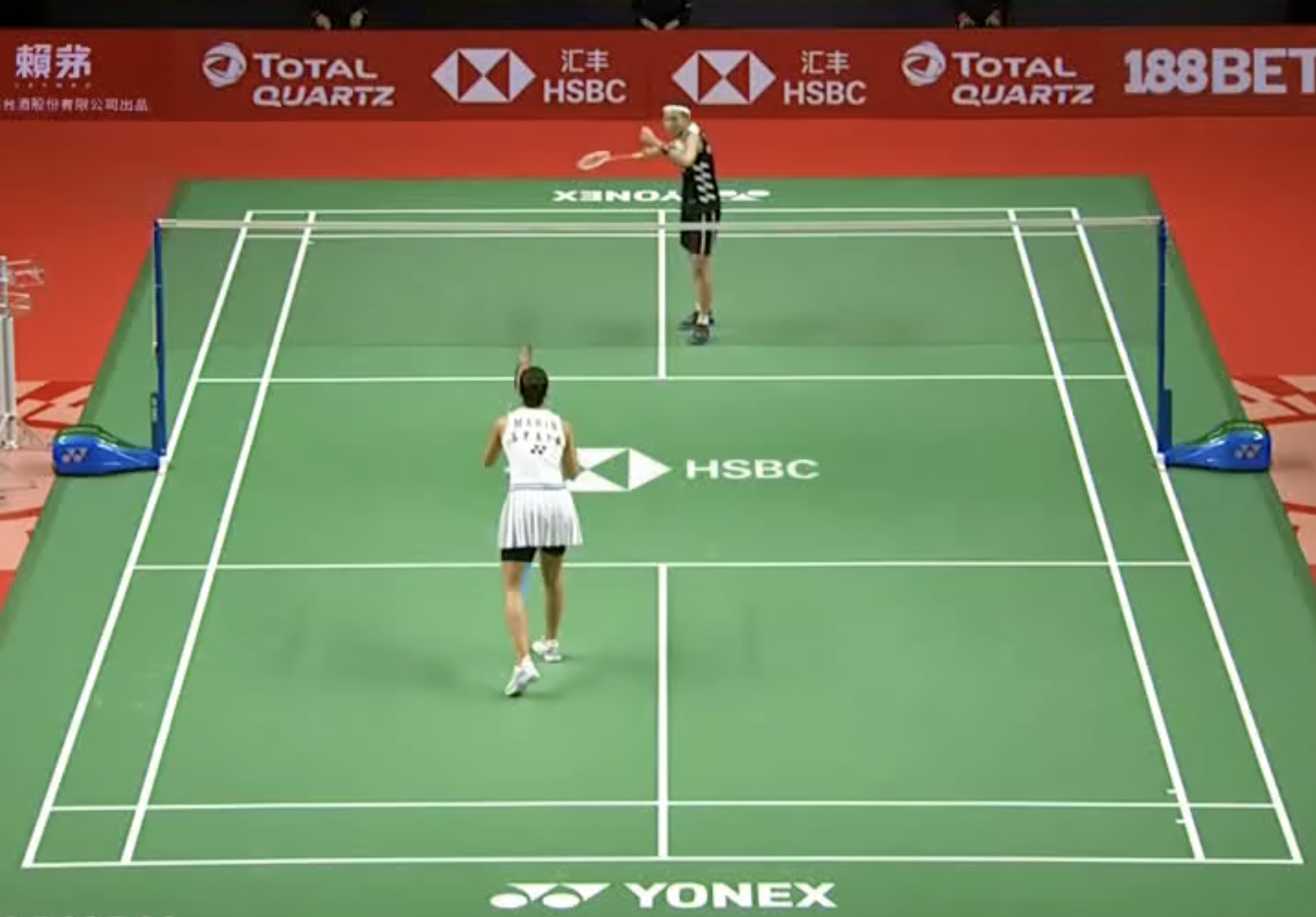}\linebreak[0]
    \includegraphics[height=0.1\textwidth, width=0.32\columnwidth]{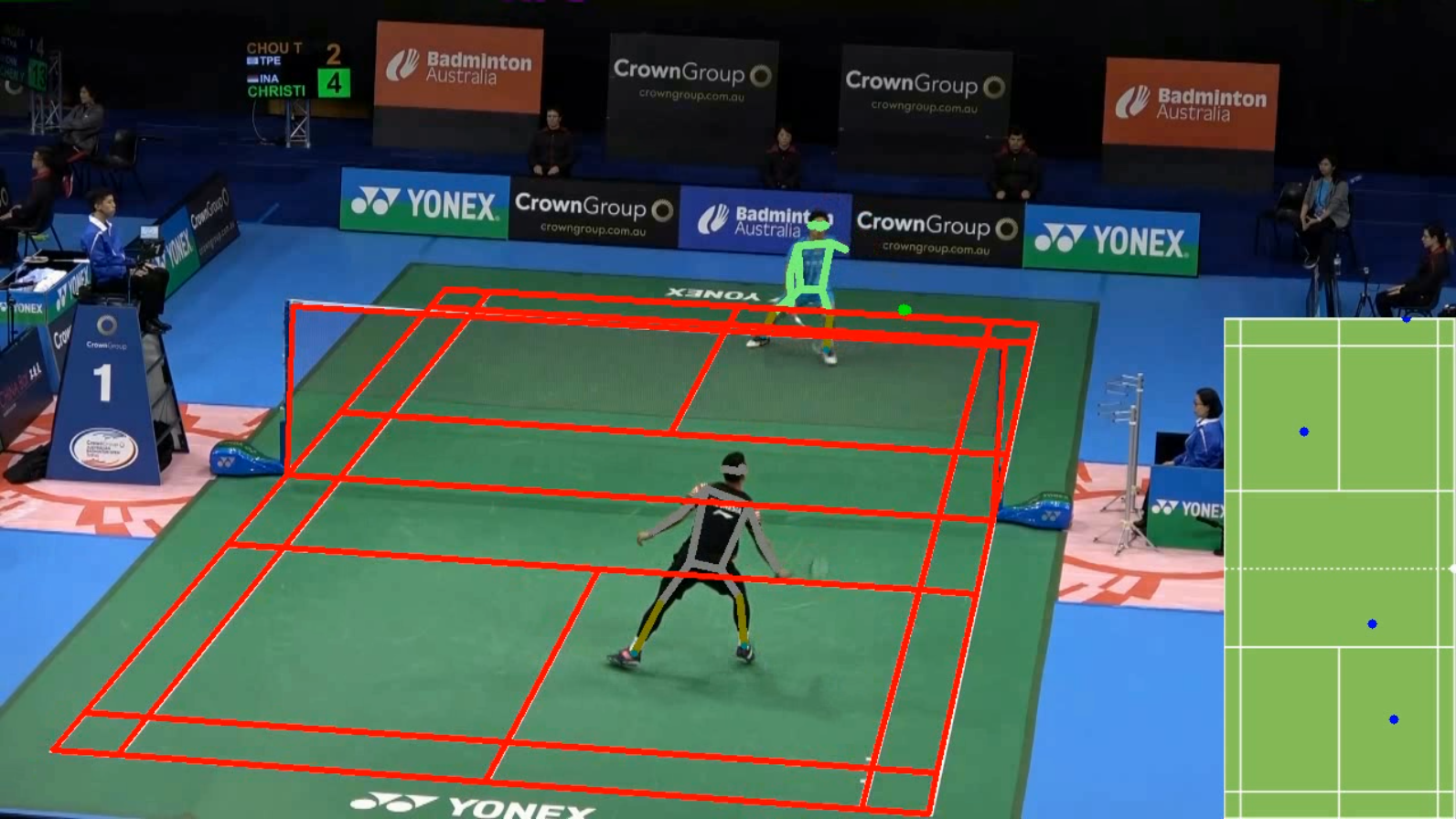}\linebreak[0]
    \includegraphics[height=0.1\textwidth, width=0.32\columnwidth]{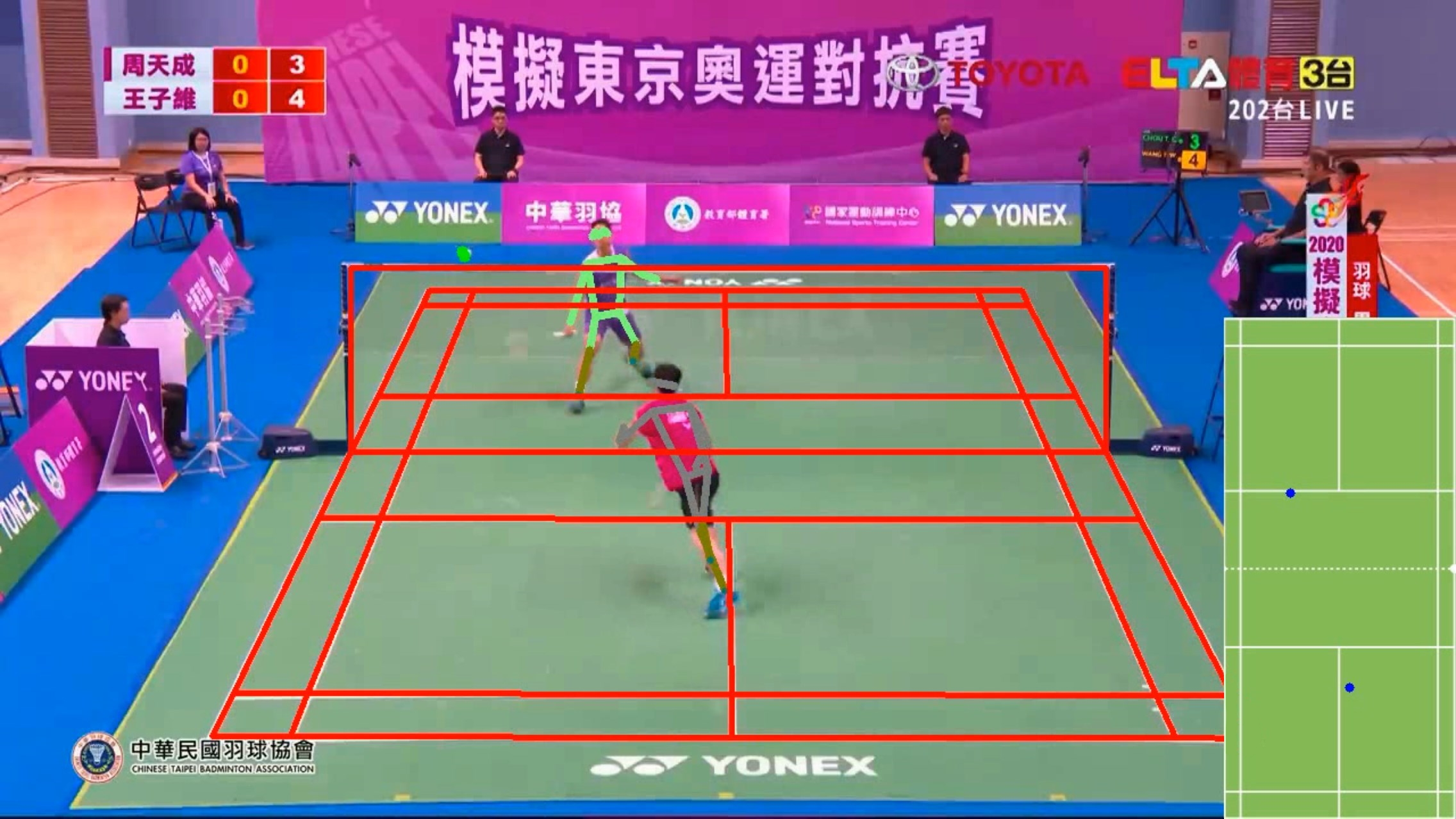}\linebreak[0]
    \includegraphics[height=0.1\textwidth, width=0.32\columnwidth]{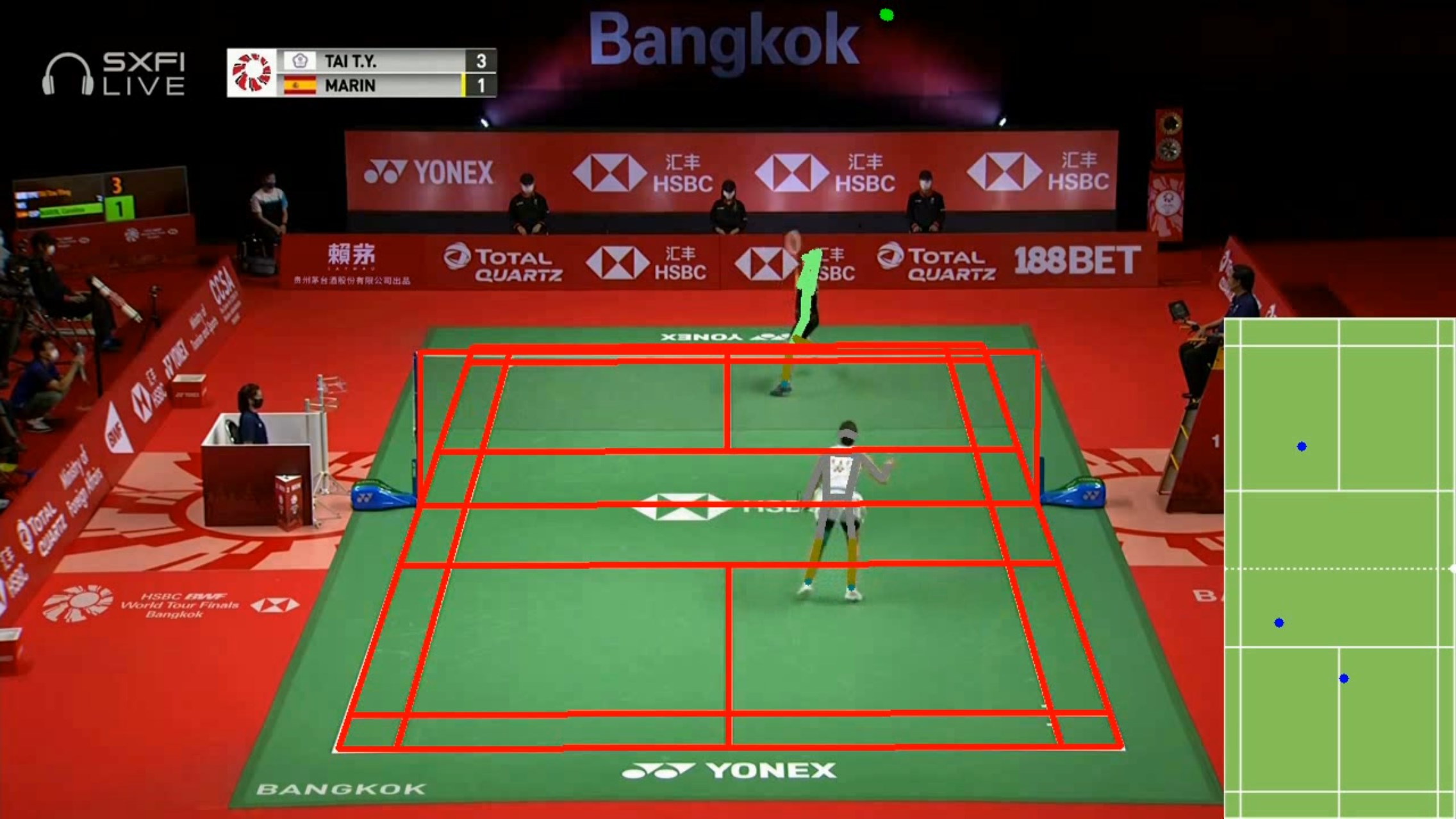}\linebreak[0]
    \includegraphics[height=0.1\textwidth, width=0.237\columnwidth]{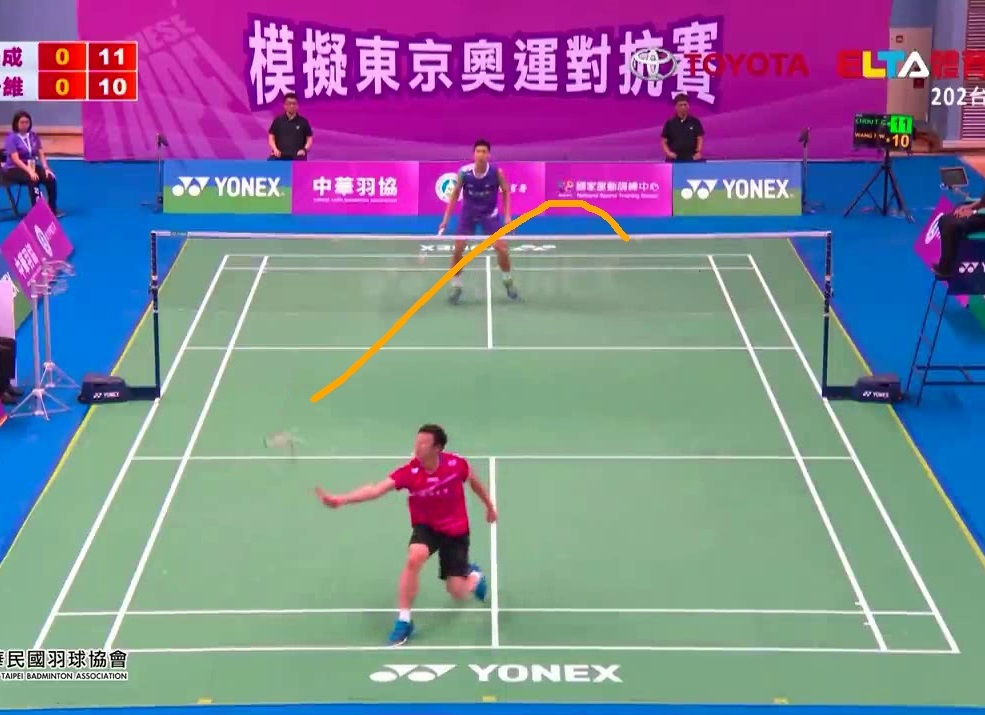}\linebreak[0]
    \includegraphics[height=0.1\textwidth, width=0.237\columnwidth]{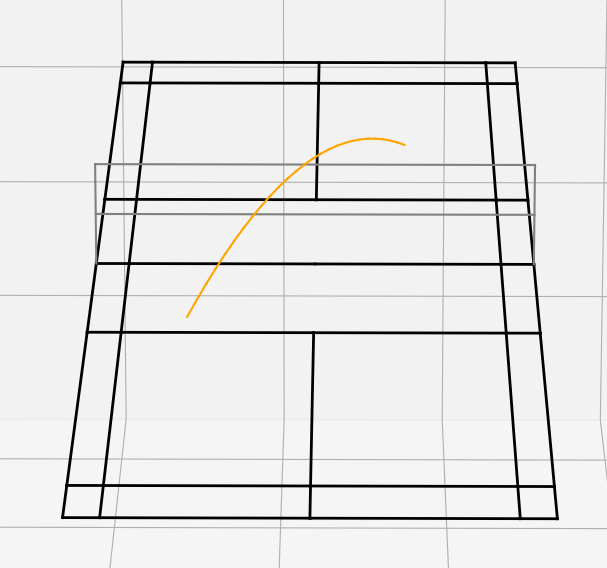}\linebreak[0]
    \includegraphics[height=0.1\textwidth, width=0.237\columnwidth]{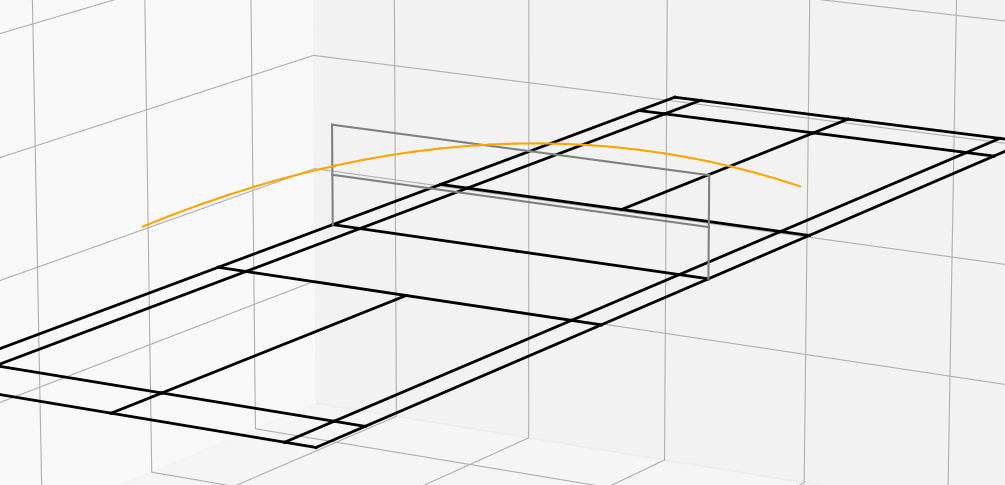}\linebreak[0]
    \includegraphics[height=0.1\textwidth, width=0.237\columnwidth]{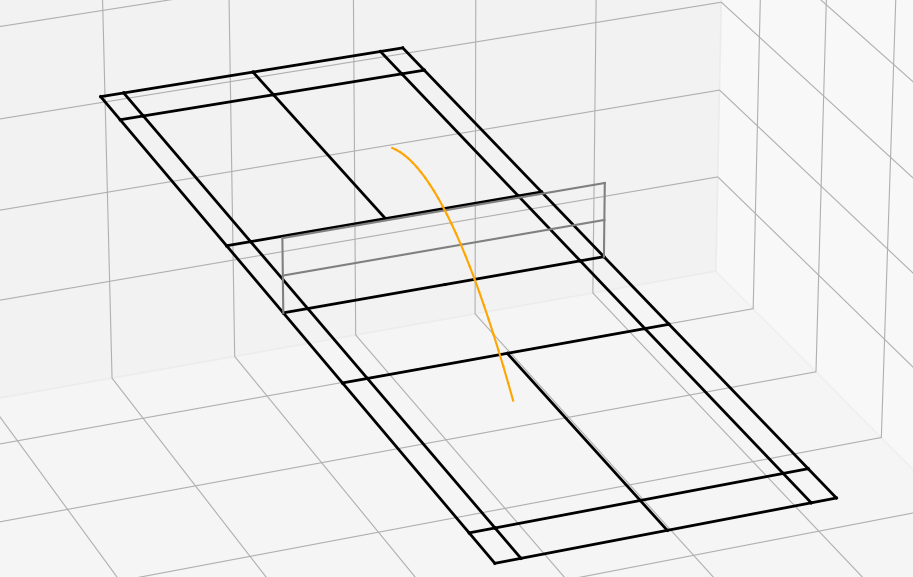}\linebreak[0]
    \caption{{\bf Court, pose, and 3D shuttlecock trajectory} automatically generated by our system. \emph{Top}: test matches used in our dataset. \emph{Middle}: court, pose, and shuttlecock frame positions inferred by our system; \emph{Bottom}: Reconstructed 3D trajectories displayed in novel camera angles.
    \label{fig:test-matches}}
\end{figure}

In a typical badminton game, spatial information such as where the player hits the shuttlecock and how the opponent responded, conveys important first order information about how the match has been progressing. Moreover, as badminton players use various height-related tactics to adjust the rhythm of the game, this spatial information is especially valuable when presented in 3D. Previous academic work, as well as our informal conversation with top-level players, confirm that the use of 3D trajectories can aid in various ways such as tactics formulation and post-game analysis~\cite{chu_tivee_2021,ye_shuttlespace_2021}.

However, recovering the 3D trajectories from monocular videos is difficult. First, in-the-wild badminton videos generally contain multiple points (\emph{rallies} in badminton terms), and each rally contains multiple \emph{shots}. Segmenting these shots out is an activity recognition problem, which can be challenging considering the extreme speed of the badminton shuttlecock.\footnote{A badminton smash can reach more than $250$ mph, which would travel across the entire court in less than half a second.} 
Even if we have the shots segmented, reconstructing the trajectory of a point (the shuttlecock) is still ill-defined due to the lack of stereo camera cues. If some reference 3D points are known, and ballistic trajectories can be assumed, then it is possible to reconstruct trajectories for each shot, such as the work done for tennis~\cite{zhang_vid2player_2020} and basketball~\cite{chen_physics-based_2009}. However, unlike these sports, the badminton shuttlecock is heavily affected by air drag, and can easily get damaged within the course of a rally. The shuttle is also not allowed to touch the ground during play, thus eliminating any useful physical information that can be inferred from the bounce of the shuttle. Other ball sports can additionally use the location of players' feet to localize the ball. However, the frequent jumping of badminton players, render many of these prior methods infeasible. To make matters worse, even 2D shuttlecock tracking is highly non-trivial: badminton is predominantly an indoor sport, with a small court and extremely fast shuttlecock speed; the shuttlecock is tiny, can be occluded by the player, and can go out of the frame frequently. Altogether, this results in limited success achieved in prior work. For example, the method proposed by~\cite{lames_reconstruction_2018,lames_measurement_2020} requires human intervention for every shot. In contrast, our system requires no human intervention and can significantly outperform the model-based work by~\cite{lee_badminton_nodate}. 

\begin{figure*}[ht]
\centering
\includegraphics[width=0.9\textwidth]{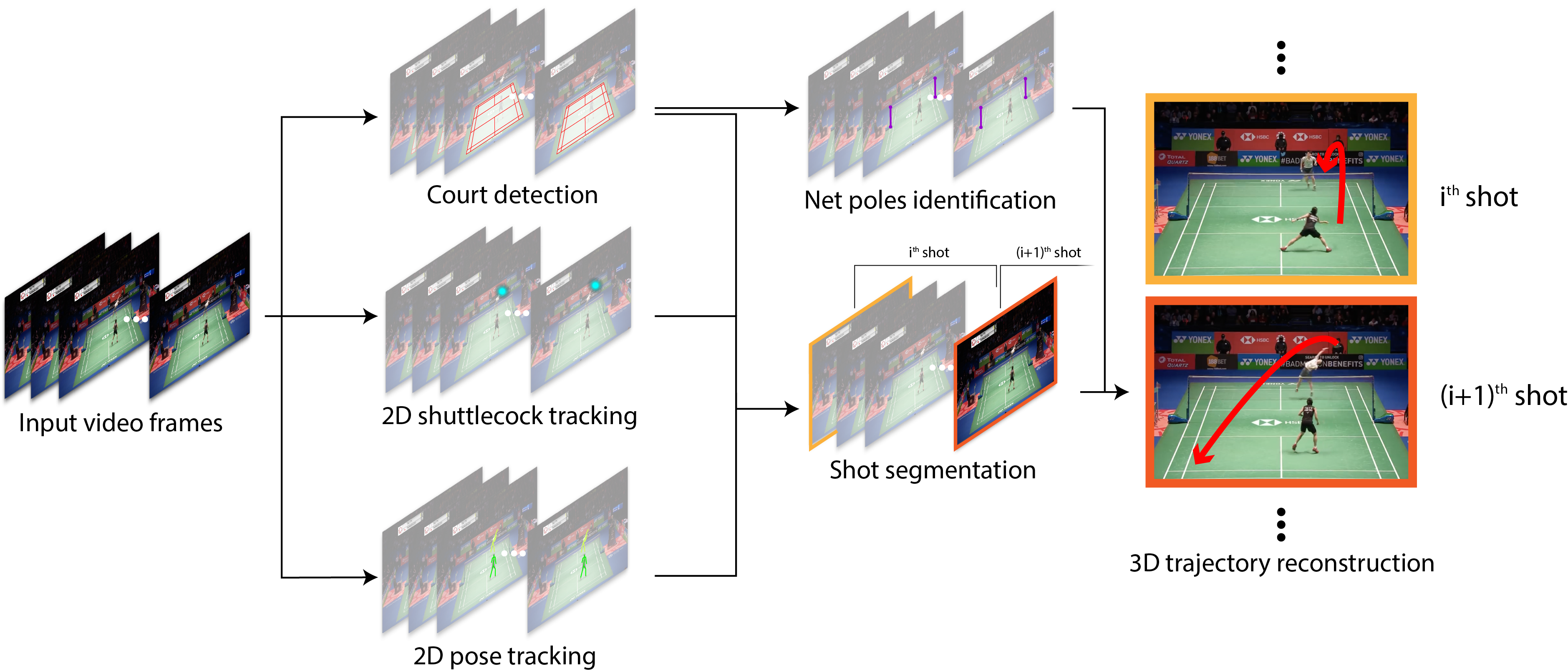}
\caption{{\bf Our method} leverages detected court, shuttlecock track, and player poses to segment a sequence of video frames into \emph{shots}, and reconstruct faithful 3D trajectory for each shot using a nonlinear optimizer.}
\label{fig:overview}
\end{figure*}

\paragraph{Our contribution} Our work tackles the holistic problem of segmenting and reconstructing the trajectory of shots from unlabeled, monocular videos. Our approach consists of a set of subsystems that analyzes a given video to identify the court, player poses, and per-frame shuttlecock pixel positions. Using these signals, we train a recurrent network to obtain the segmentation of shots. Our network is compact, efficient to train, and results in high accuracy in detecting the shots (See Table~\ref{tab:hit-detection}). We then propose a novel per-shot trajectory reconstruction method that leverages nonlinear optimization and domain knowledge when possible. We evaluate the 3D reconstruction method with a dataset containing real and synthetic trajectories and show state-of-the-art performance. As a simplification, we limit the scope in this work to singles play, where only one player per side of the court is allowed. We hope that the statistics provided by this system can enable players to achieve greater levels of play through more efficient data-based tactical analysis. \Cref{fig:overview} provides an overview of our system. 

\section{Related Work}
\label{sec:rel}


Prior work has generally focused on specific components such as court detection~\cite{kim_soccer_2000,yamada_tracking_2002,FKWE04,watanabe_soccer_2004,wang_fast_2006,homayounfar_sports_2017}, activity localization and classification~\cite{cioppa_bottom-up_2018,cioppa_context-aware_2020,tsunoda_football_2017,schwarcz_spin_2019}, stroke analysis~\cite{kulkarni_table_2021}, pose analysis~\cite{hong_video_nodate}, or ball tracking~\cite{huang_tracknet_2019,sun_tracknetv2_2020,sarkar_generation_2019,theagarajan_soccer_2018}. Our system, on the other hand, aim to generate and integrate different vision-based sub-signals including court, pose, shuttlecock positions to segment unlabeled videos into the known cyclic structure of point rallies~\cite{zhang_vid2player_2020}, and produce 3D trajectories of the shuttlecock that can, for example, be used in advance visualizations to improve tactics selection~\cite{chu_tivee_2021,ye_shuttlespace_2021}. Starting with some of these baseline sub-signal models, we have identified a number of key improvements to each model to boost the overall performance of our system, detailed in \S\ref{sec:annotation_pipeline}.

3D reconstruction is a widely studied topic in vision~\cite{szeliski_computer_2010}, however, its use in sports videos is relatively recent. \cite{lames_reconstruction_2018,lames_measurement_2020} showed a confirming point method that can reconstruct 3D shuttlecock positions but requires human intervention in placing the confirming points for every shot. ShuttleSpace showed that 3D trajectories of badminton shots is useful for top-level players in an immersive analytics system~\cite{ye_shuttlespace_2021}. TIVEE further confirmed that 3D trajectories conveys important tactical information and can be used to improve game planning and post-game analysis~\cite{chu_tivee_2021}. Both ShuttleSpace and TIVEE were based on confirming points and thus requires human intervention at shot level to get accurate 3D trajectories. Vid2Player showed that 3D trajectories of tennis ball can be used to substitute heavily motion-blurred broadcast footage to train a player behavioral model for synthesis~\cite{zhang_vid2player_2020}. \cite{chen_physics-based_2009} used 3D trajectories of basketball shooting to infer shooting location statistics. Vid2Player used manually annotated shot boundaries, and \cite{chen_physics-based_2009} is based on shot boundary detection using histogram descriptors. Finally, \cite{lee_badminton_nodate} showed a model-based trajectory estimation method based on linear regression and SVM. However, their method is only evaluated on 2D synthetic shot trajectories. To our knowledge, our work is the first end-to-end system that can provide shot segmentation and 3D shot trajectory estimation \emph{without human intervention}. 

\section{Dataset}
\label{sec:dataset}

Our dataset is based on the public TrackNetV2 dataset~\cite{sun_tracknetv2_2020,huang_tracknet_2019}. In total, this dataset contains 77k annotated frames from 26 unique singles matches from international play, filmed from an overhead, static, ``broadcast-view" camera. Following the approach of TrackNet, we use the 12k frames from their three ``test matches" for testing, and split the rest with 10\% of the frames for validation and 90\% for training. The dataset also contains timestamp information of when either player stroke the shuttlecock (a \emph{hit}). We enhanced this dataset by labeling the four court corners for each match, and identify the player who hit the shuttlecock whenever hit is present.

Since the TrackNet dataset contains mostly ``broadcast views" (where the camera is situated in the bleachers behind the court), we additionally labeled another 40 matches to test our court detection algorithm. These additional matches were mined from YouTube and filmed from closer angles that were not present in the TrackNet data. As described in \Cref{sec:experiments}, we also prepare a synthetic dataset that contains 10k 3D trajectories simulated from a physical model.
\begin{figure}[ht]
    \centering
    \includegraphics[width=0.49\columnwidth]{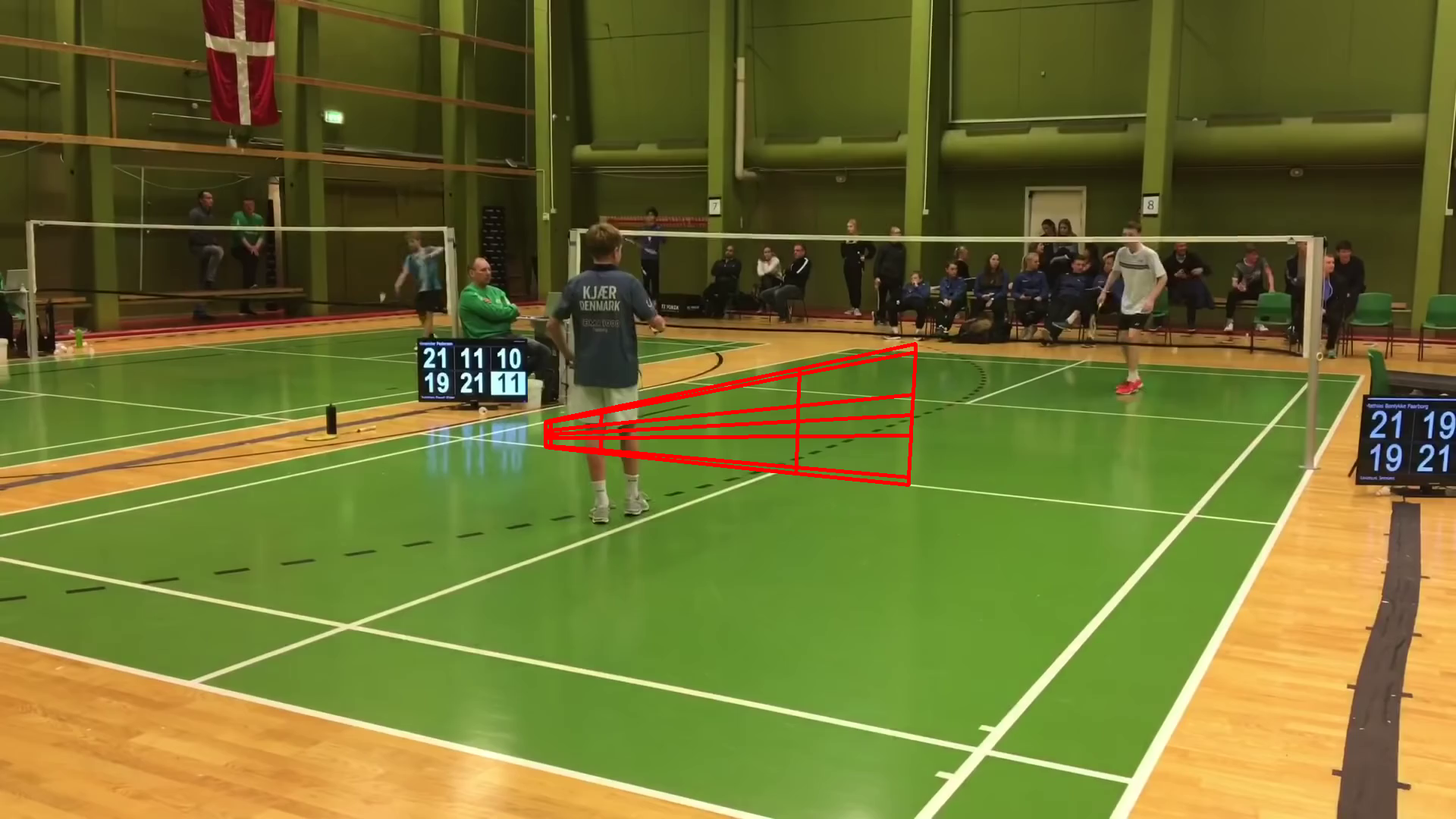}\linebreak[0]
    \includegraphics[width=0.49\columnwidth]{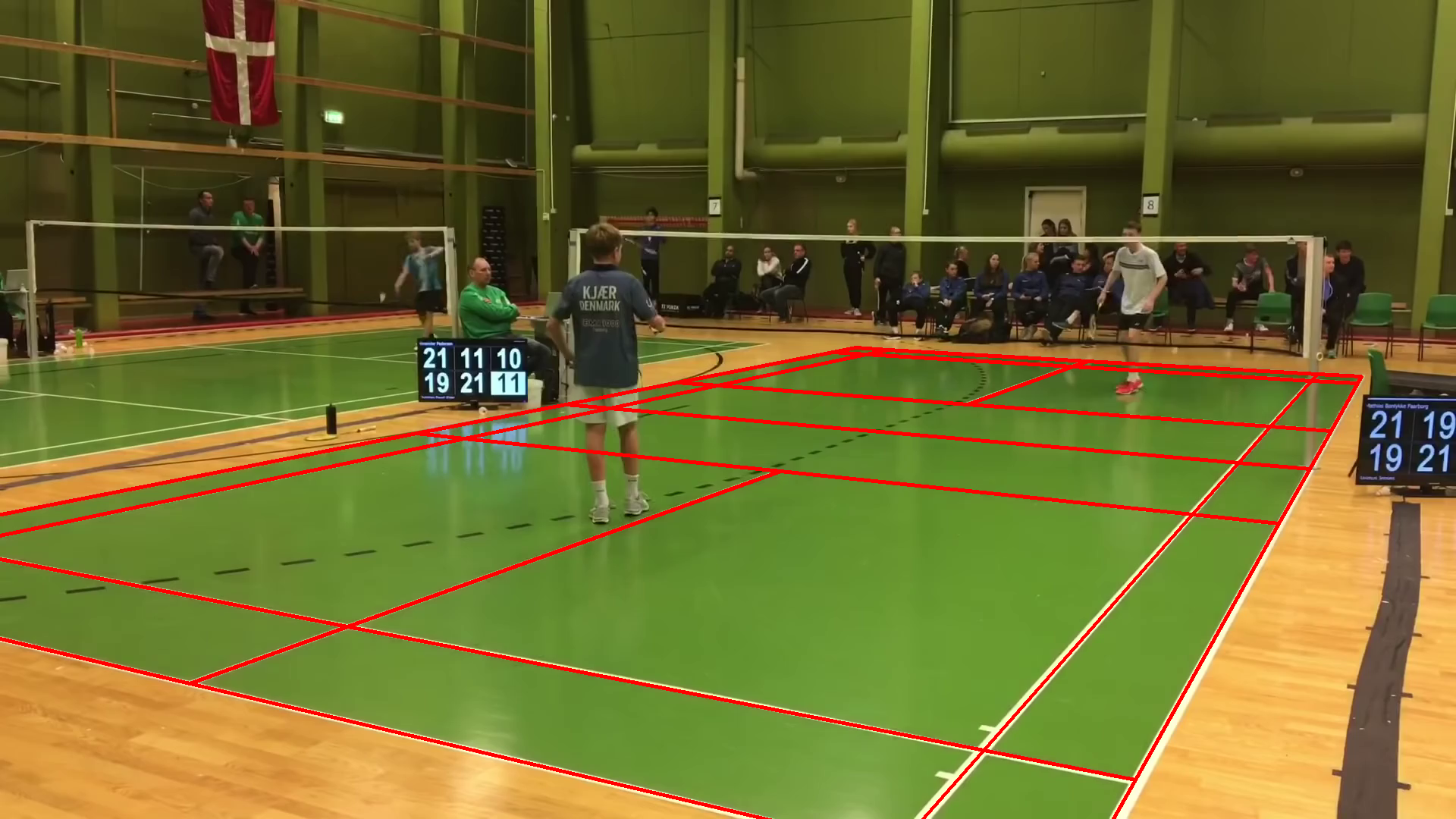}\linebreak[0]
    \includegraphics[width=0.49\columnwidth]{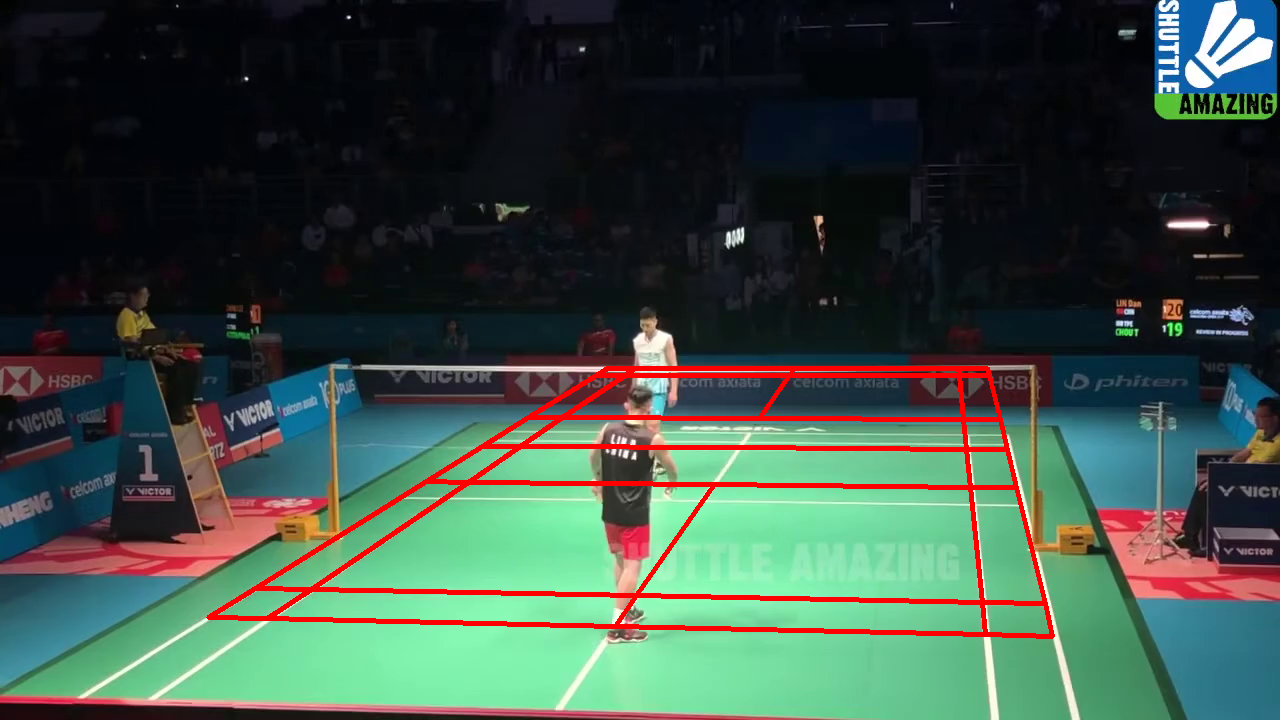}\linebreak[0]
    \includegraphics[width=0.49\columnwidth]{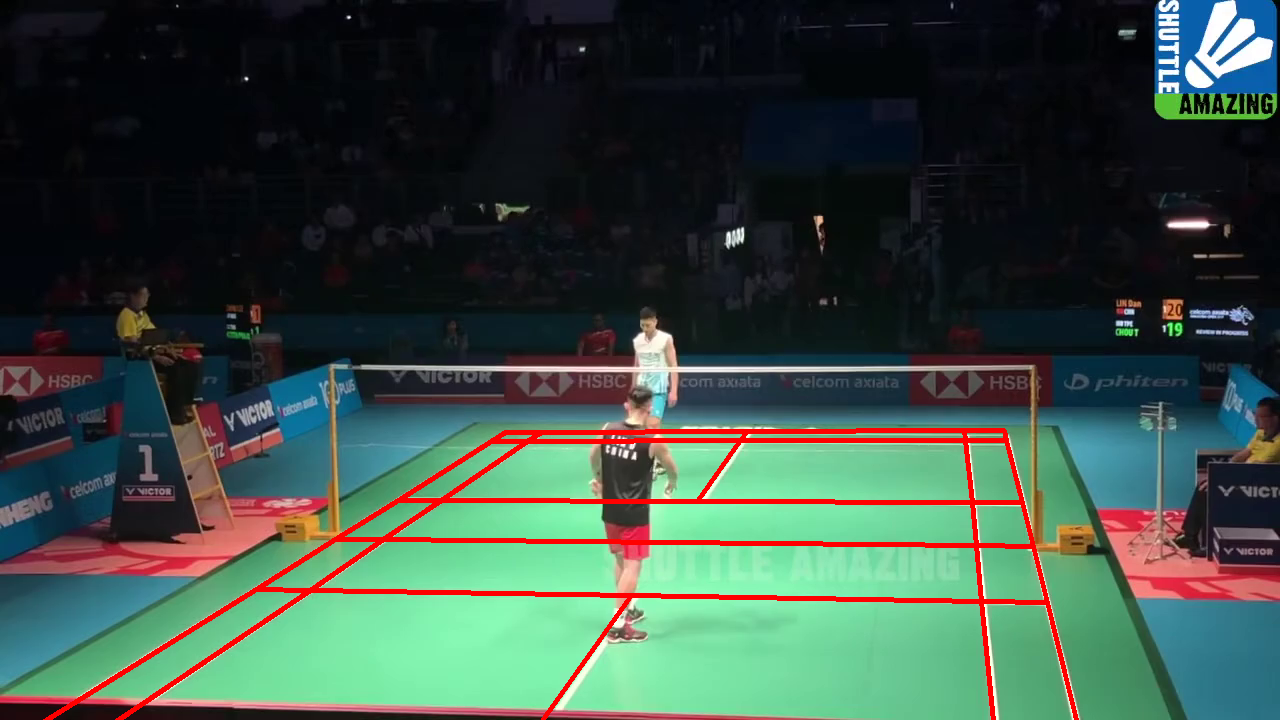}\linebreak[0]
    \caption{{\bf Our court detection model is more robust} than the baseline Farin \etal method~\cite{FKWE04}. \emph{Left}: Courts detected with \cite{FKWE04}. \emph{Right}: Courts detected with our model. These two images are from additional annotated matches outside of the TrackNet dataset.
    \label{fig:court-improvements}}
\end{figure}

\section{Automated 3D trajectory reconstruction}
\label{sec:annotation_pipeline}
In this section, we present each part of our system in detail. For each part of the system, we review the existing state-of-the-art in the area, and discuss modification specific to our work where appropriate. As previously described, our system currently only analyzes singles rallies recorded with a fixed camera from an approximate ``broadcast view" (see~\Cref{fig:test-matches} for examples of this view).

\subsection{Court detection}
\label{sec:court-detection}

We base our approach for detecting the court on the model-based algorithm of Farin \etal~\cite{FKWE04,chen_physics-based_2009,LLHG06,CTLY12}. Unfortuntately, we found that the original algorithm fails on many videos in our dataset where the viewing angle is off-center. We propose a graph-based approach to overcome this issue and significantly boost the performance of this algorithm.

\begin{figure*}[ht]
\includegraphics[height=0.13\textwidth]{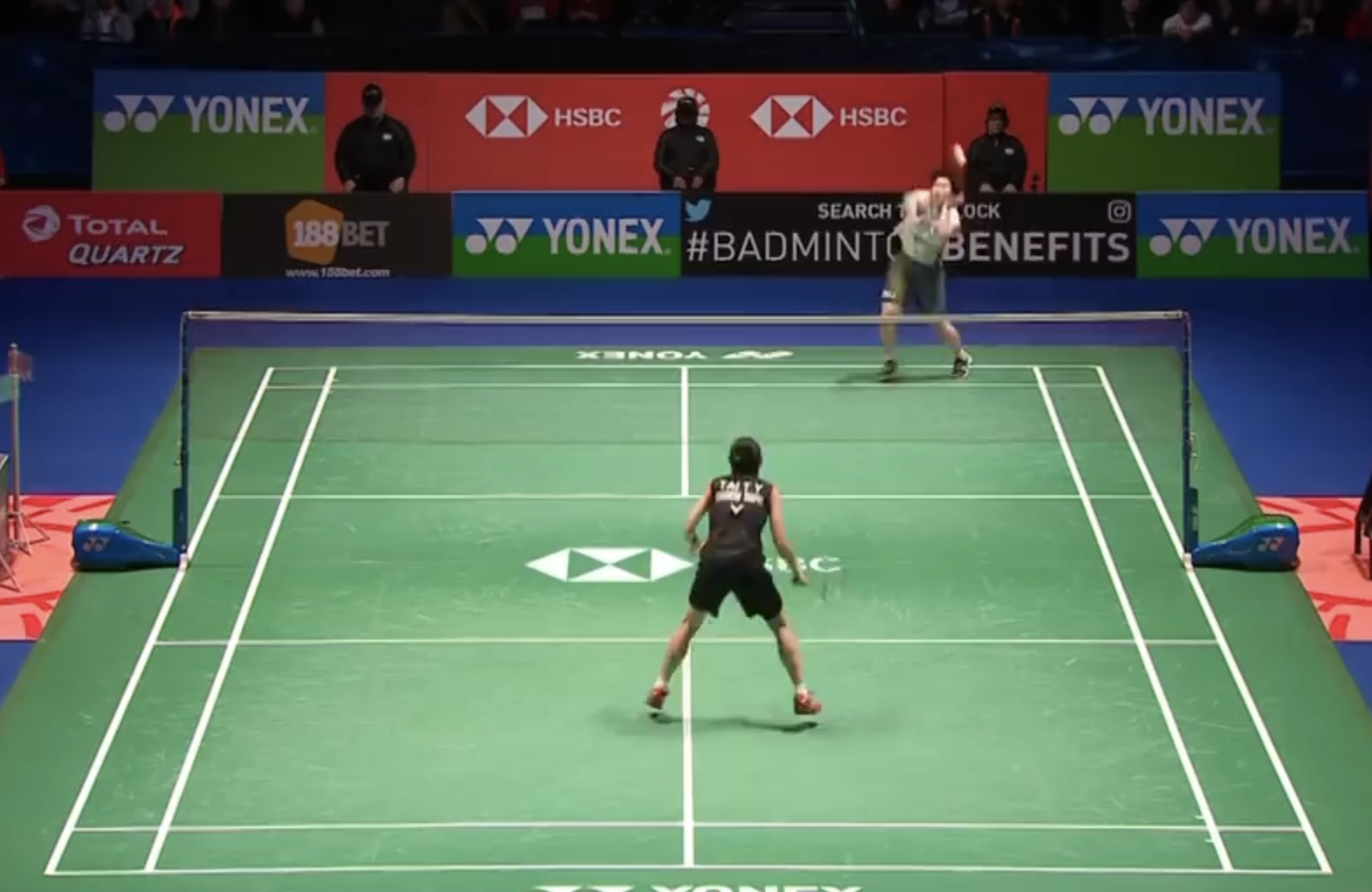}\linebreak[0]
\includegraphics[height=0.13\textwidth]{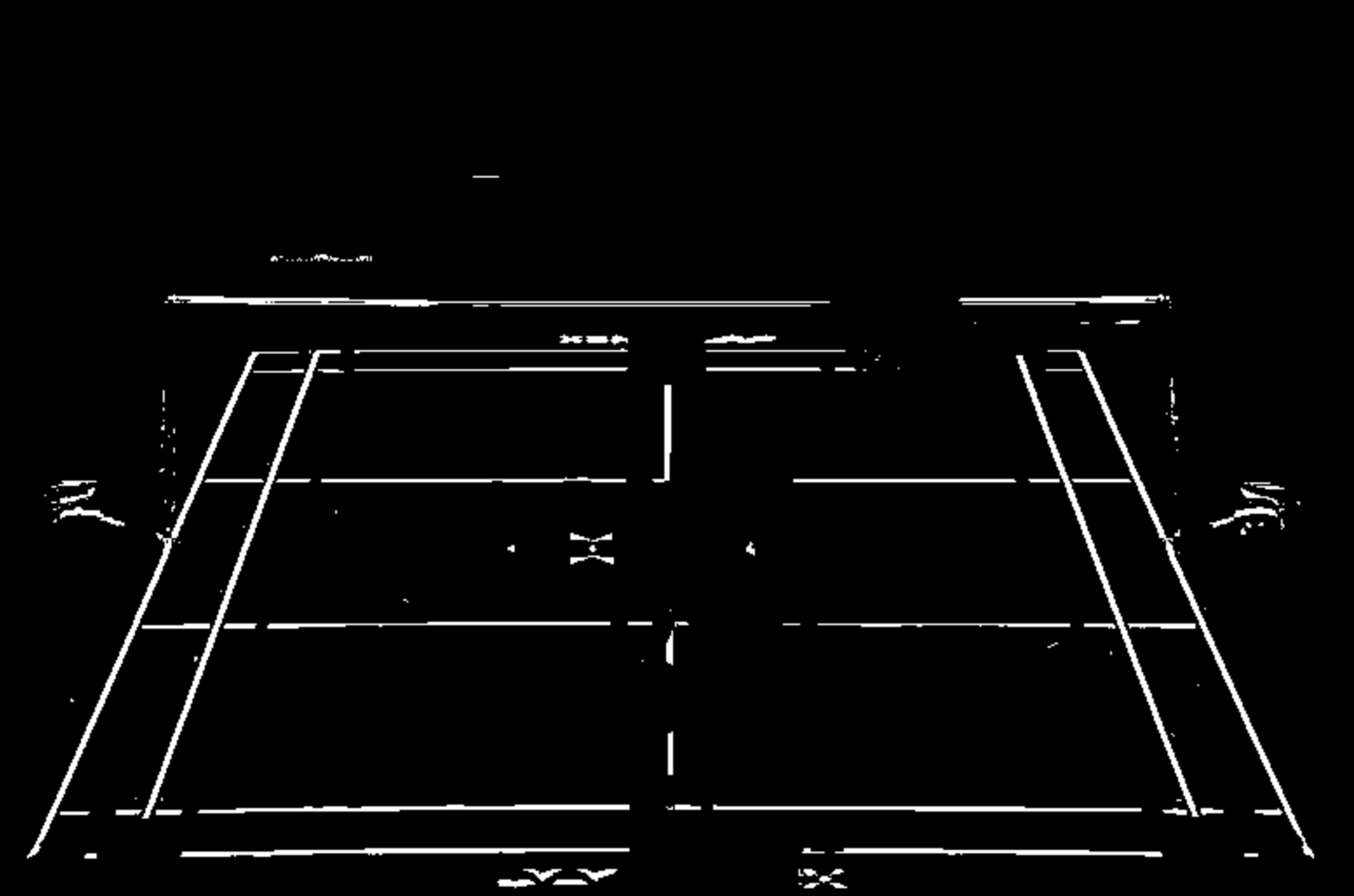}\linebreak[0]
\includegraphics[height=0.13\textwidth]{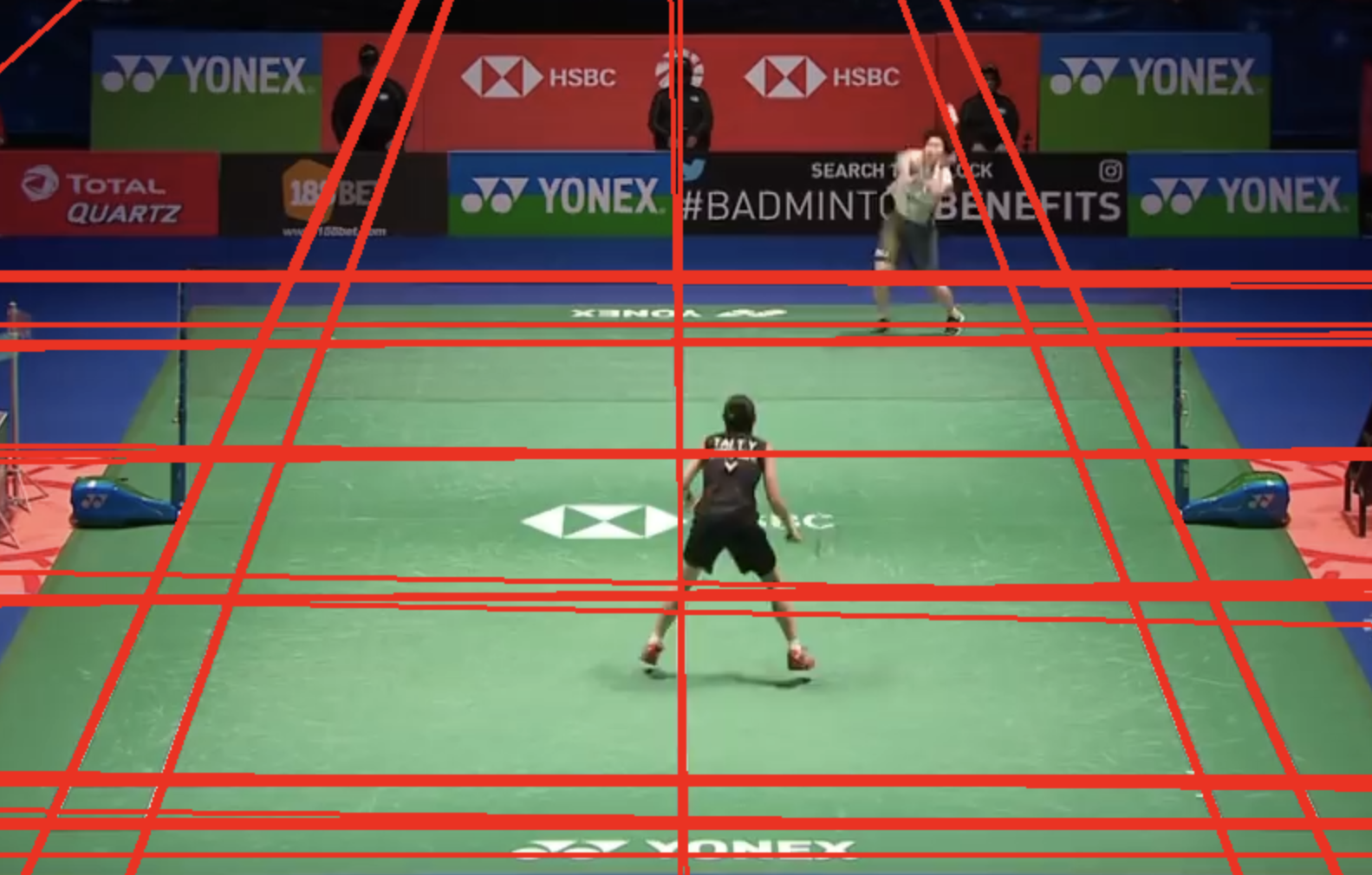}\linebreak[0]
\includegraphics[height=0.13\textwidth]{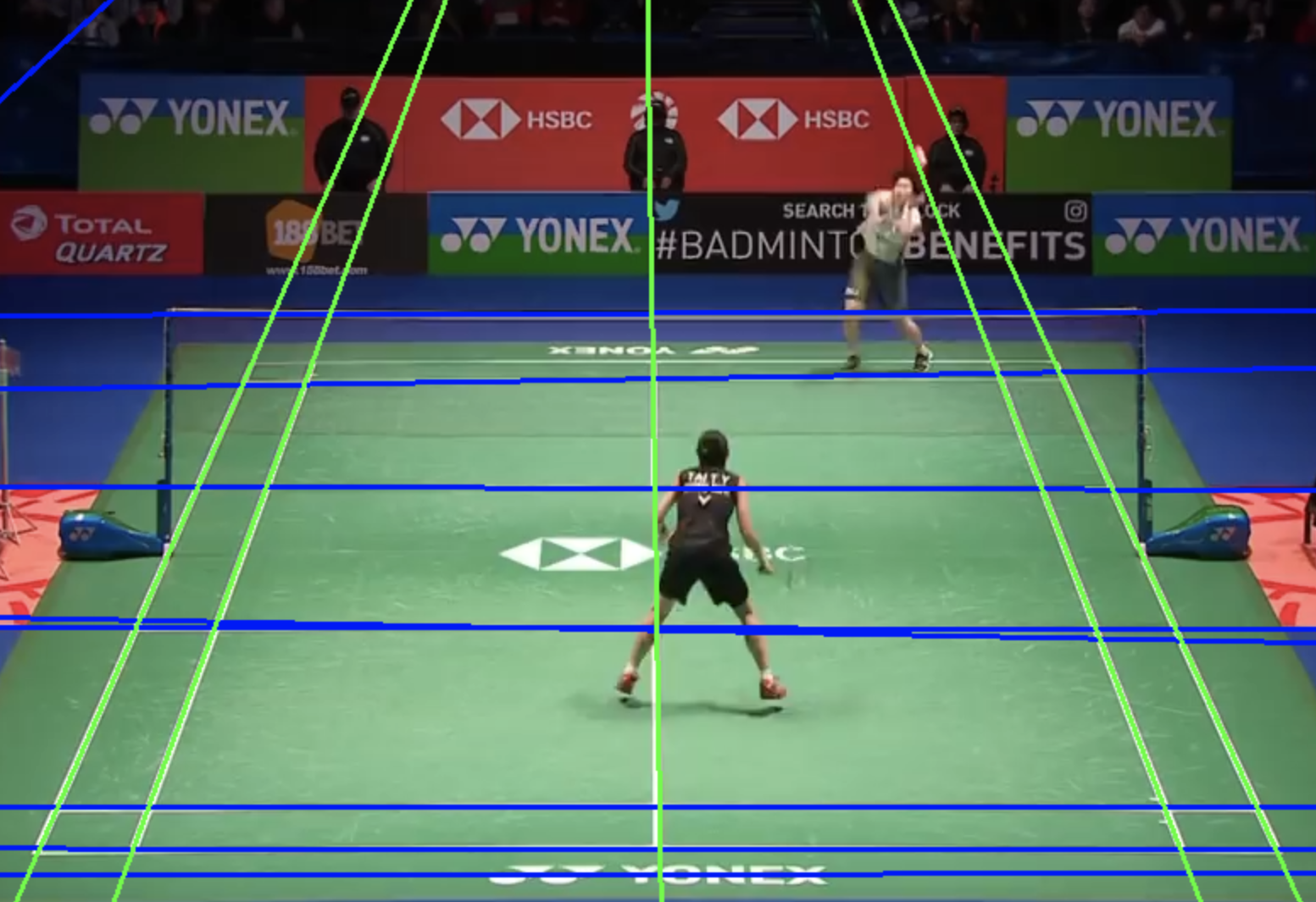}\linebreak[0]
\includegraphics[height=0.13\textwidth]{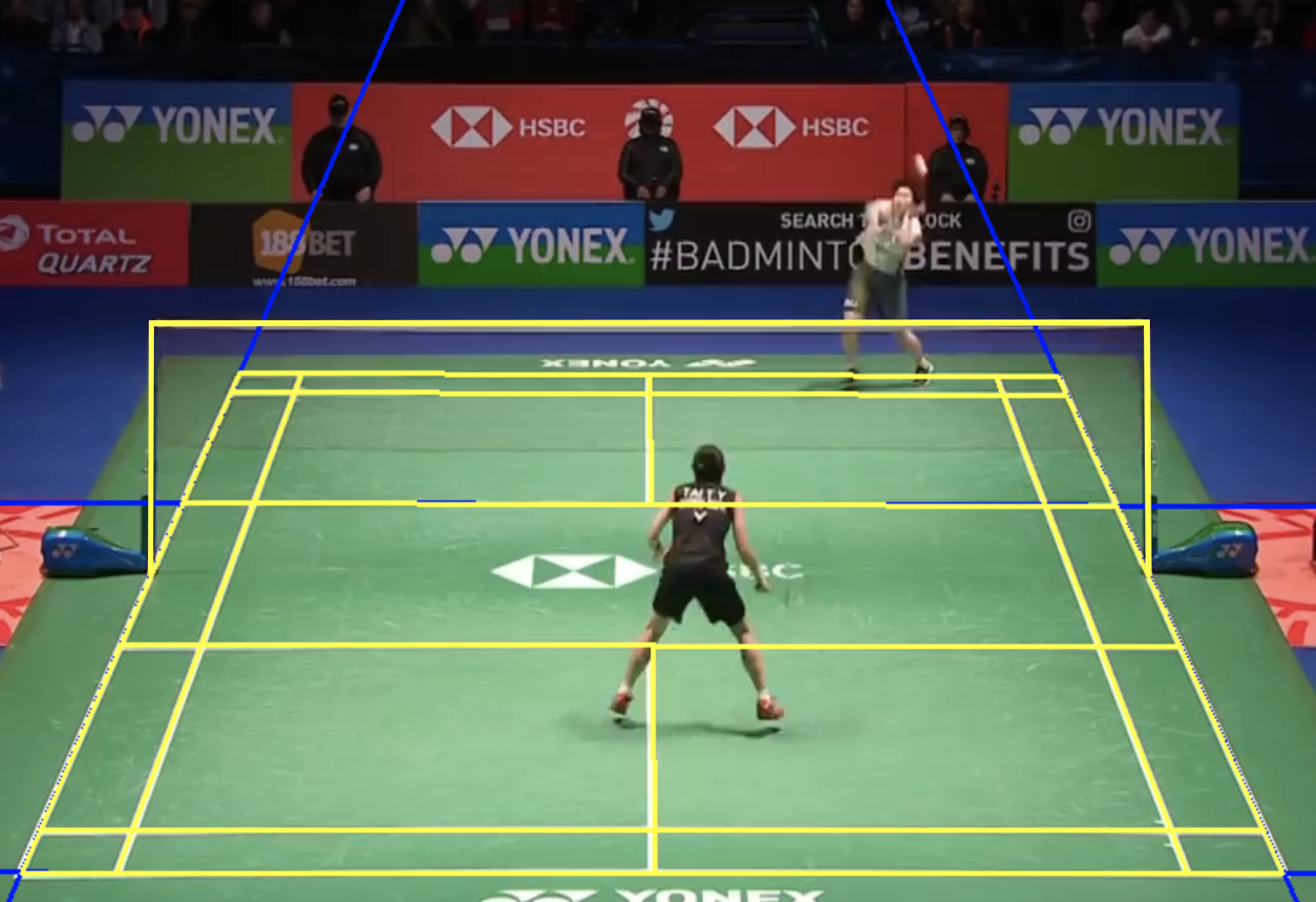}\linebreak[0]
\caption{{\bf Stages of our court detection} includes applying pixel color thresholding (second) and a Hough transform to obtain line candidates (third), and then partitioning these lines (fourth) in order to efficiently search for the correct court layout (last). \label{fig:court-detect}}
\end{figure*}

We briefly illustrate the original Farin's algorithm~\cite{FKWE04} before introducing our modification. Given an image of the court, candidate lines are first detected using pixel color thresholding followed by application of the Hough transform (see~\Cref{fig:court-detect}). The detected lines are then split into a set $L_H$ of horizontal lines (those with slopes between -25 and 25 degrees) and $L_V$ of vertical lines (those with slopes between 60 and 120 degrees). Finally, a combinatorial search is conducted to match a known court layout to the candidate lines, with each search iteration picking two horizontal and two vertical lines for a reference rectangle in the known layout. This results in a $\mathcal{O}(|L_H|^2|L_V|^2)$ algorithm. Unfortunately, this algorithm fails in about 24\% of videos in our dataset. We found the main culprit to be the hard angle constraints set when partitioning results in line misclassification, which ultimately break the algorithm. This happens most frequently when the angle of the camera is not ideal or when the reference rectangle of the court is not visible (see~\Cref{fig:court-improvements} for examples).

To avoid this issue, we propose a new partitioning algorithm that is free of hard-coded angle constraints. To do this, we frame the problem as a maximum weight bipartite subgraph identification. We represent each line as a node in a complete graph, and for every pair of lines $u$ and $v$, we connect the nodes with an edge of weight $(|\mathrm{angle}(u, v) - \pi / 2|+ \varepsilon)^{-2}$ ($\varepsilon$ is set to be a small constant, e.g. $10^{-2}$), where $\mathrm{angle}(u, v)$ is the angle between lines $u$ and $v$ closest to $\pi / 2$. Then we greedily try to partition the graph into two sets of vertices $L_H$ and $L_V$ such that the weight between the two sets is maximized. This weight function encourages a partition where the lines in the two parts are roughly orthogonal to each other.

Our implementation is developed on top of the reference code provided by~\cite{court_detect}. To assess the accuracy of the court detection, we measure two metrics over a manually annotated dataset: (i) the average pixel error over all detections, and (ii) the percentage of successful detections. We define a court detection as successful if the IoU of the detected court and ground truth is $>0.8$. On our dataset, our proposed approach increases the success rate of court detection from 73.9\% to 85.5\%, and decreased the average detection time by a factor of 40 while achieving an higher average IoU of 0.97 (vs. 0.96 from the original method).

\subsection{Pose estimation}
\label{sec:pose-estimation}
We perform pose estimation using a top-down HRNet model~\cite{hrnet} to compute per-frame poses through the {\tt mmpose} framework~\cite{mmpose}. To track poses, instead of using methods developed for unstructured environments~\cite{girdhar_detect-and-track_2018} or recurrent network-based methods~\cite{sadeghian_tracking_2017}, we simply leverage the detected court as a strong cue. We filter all detected poses that do not have feet in the court, and identify near and far players based on their distance to the camera. This strategy is effective due to the fact that no one other than the players can step onto the court, and that players do not switch side during a point. To accommodate jumping motions, which would misplace the player to a deeper position than they actually are, we make two modifications to increase robustness. Firstly, we relax the court boundary slightly. Secondly, if a side of the court has no pose within it, we find the pose closest to the last in-court pose recorded on that side. For all of our videos, this simple approach identifies the two players on every frame.


\subsection{Shuttle detection}
\label{sec:shuttle-tracking}
\begin{figure}[ht]
    \centering
    \includegraphics[width=\columnwidth]{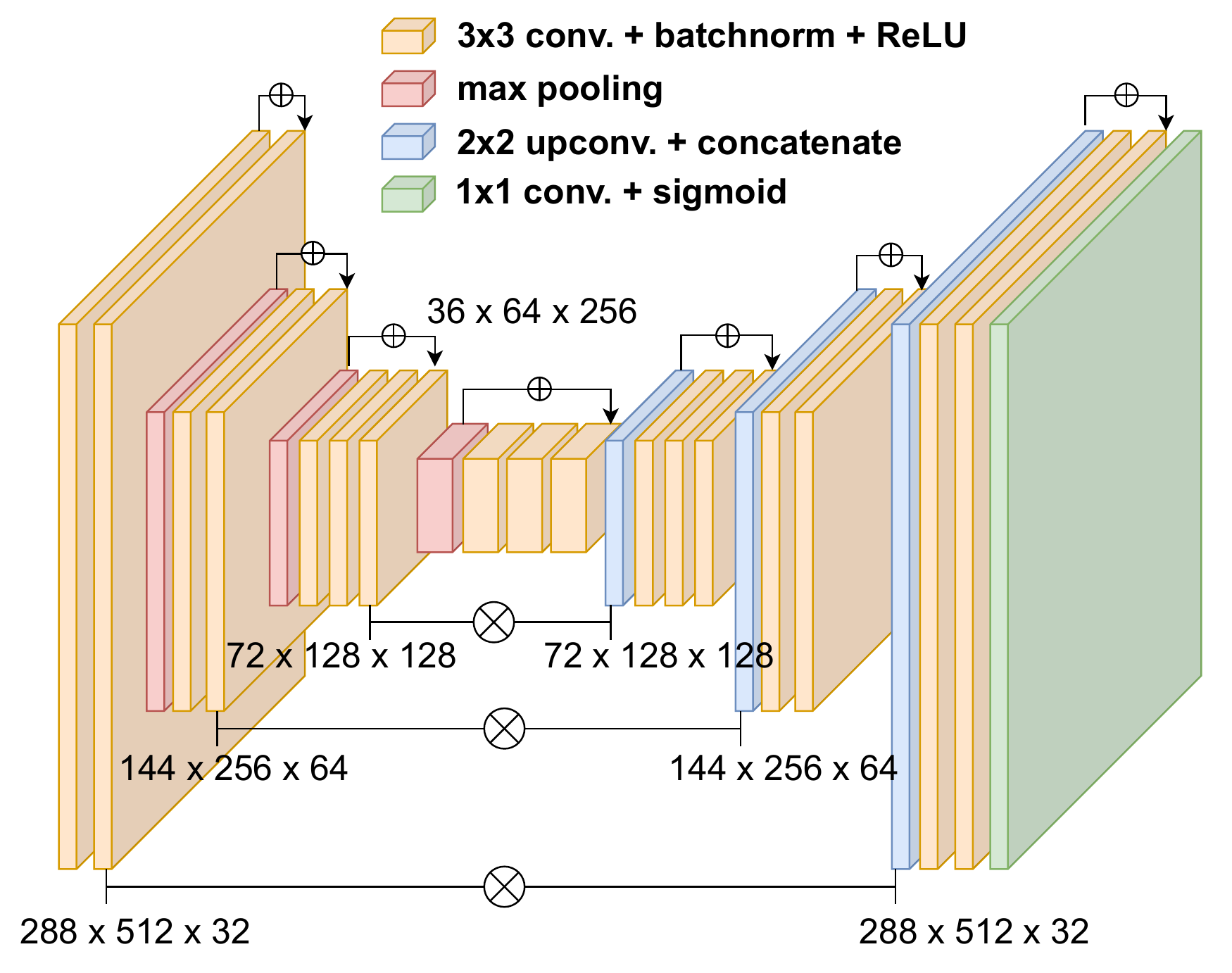}
    \caption{{\bf Architecture of our shuttle detection model} is based on a modified U-net, where we added residual connections and use the weighted dice and binary cross-entropy loss. $\bigoplus$ represents addition and $\bigotimes$ represents concatenation.}
    \label{fig:our-tracknet}
\end{figure}

We formulate the shuttle detection problem as semantic segmentation and use a U-net style architecture (\Cref{fig:our-tracknet}) inspired by TrackNetV2~\cite{sun_tracknetv2_2020}. Similar to TrackNetV2, to encourage the network to learn the temporal context, we use a 3-in-3-out architecture that predicts the shuttle masks for three consecutive frames simultaneously (each resized to 288-by-512). However, our model has significantly smaller footprint than the original TrackNetV2 (2.9M parameters in our model versus 11.3M parameters), making it faster to train and perform inference, and higher accuracy ($88.6\%$ accuracy from $84.0\%$ in the original). This improvement is credited to two main changes we introduced. First, we added residual connections to each convolutional layers (see~\Cref{fig:our-tracknet}). Second, instead of a binary focal loss, we use a weighted combination of the dice loss and the binary cross-entropy to mitigate the input imbalance problem of tiny shuttlecock, inspired by \cite{combo_loss}. Given per-pixel prediction $\hat{y} \in [0,1]$ and ground truth label $y$, the loss $L$ we use is
\begin{align*}
    L_{B}(y, \hat{y}) &:= y^T \log(\hat{y}) + (\mathbf{1}-y)^T \log (\mathbf{1} - \hat{y}), \\
    L_{D}(y, \hat{y}) &:= \frac{y^T \hat{y}}{||y||_1 + ||\hat{y}||_1 + \varepsilon}, \\
    L(y, \hat{y}) &:= (1 - \alpha) L_{B}(y, \hat{y}) + \alpha L_{D}(y, \hat{y})
\end{align*}
where $\alpha$ is the blending coefficient, and $\varepsilon$ is a small constant for numerical stability. Throughout our experiments, we use $\alpha=0.1$ and $\varepsilon=10^{-4}$. To generate the final shuttlecock location, we threshold $\hat{y}$ at $0.5$ to produce a binary mask per frame, and then use the centroid of the largest connected component in the mask. If no pixels are above $0.5$ or the area of the component is not large enough, we report the shuttle is undetected.

To improve training, we trained the first few epochs of our network with distillation learning~\cite{HVD15} using parameters from TrackNetV2. In total we used 10 distillation epochs and 40 training epochs with an Adadelta optimizer. Standard augmentation such as random rotations, shears, and zooms were applied.

\subsection{Shot segmentation}
\label{sec:shot_segmentation}
\begin{figure}[ht]
    \centering
    \includegraphics[trim={0cm, 0.75cm, 0cm, 0.75cm}, clip, page=2,width=0.7\columnwidth]{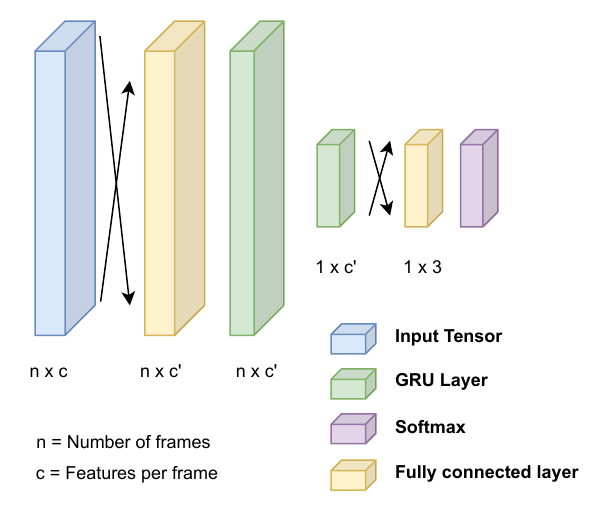}
    \caption{{\bf Architecture of our hit detection model} is based on a simple GRU-based recurrent network that consumes court, pose, and 2D shuttlecock information to make hit predictions.}
    \label{fig:hitnet}
\end{figure}

Identifying the shots of a rally is critical for reconstructing 3D trajectories (\S\ref{sec:3d-recon}) and other downstream applications. A shot happens when a player hits the shuttle with her racket, and ends right before the opposing player hits the shuttle or if the shuttle hits the court. Therefore, in order the segment the video into successive shots, it is equivalent to identify the \emph{hit} events.

We formulate the hit detection as a multi-class classification problem. Since we are focusing on singles matches only, this is a three way classification with labels \emph{no hit}, \emph{near player hit}, and \emph{far player hit} predicted at each frame.

\paragraph{HitNet: Hit detection architecture} In all racket sports, hits at certain parts of the court (e.g. the side lines) occurs more often than the others. Moreover, due to the need to efficiently translate power to the ball, athletes have very consistent poses when hitting, and have to perfect their positioning with respect to the ball. The higher the level of the athlete, the higher this consistency. As a result, we hypothesize that the court layout, the poses, and the shuttlecock location play an important role in identifying hits.

We propose a recurrent model that leverages the previously computed court layout (\S\ref{sec:court-detection}), poses (\S\ref{sec:pose-estimation}), and shuttlecock positions (\S\ref{sec:shuttle-tracking}) and their temporal tracks to predict hits. For each frame of the video, we create a feature vector comprising of the pixel coordinates of the court corners, the two players' poses, and the location of the shuttle, and normalize them jointly in the $x$ and $y$ directions. The features are embedded into 32 dimensions using a fully-connected layer, before feeding into the recurrent unit. The recurrent unit comprises of two GRU layers that takes 12 frames at a time, and predict whether a hit occurred between frame 7 to 12. We then feed the last token embedding of this recurrent unit into a small fully connected layer before passing it to the softmax layer to generate confidence scores. The architecture is shown in~\Cref{fig:hitnet}. The network is compact in size (around 16K parameters) and can reach $86.3\%$ accuracy. The network can performance inference over several thousand frames a second. To show that each feature (court, pose, shuttle) contributes significantly to the output accuracy, we perform ablation studies in \Cref{tab:hit-detection}. 

The training is done on our dataset with standard data augmentation. Artificial noise is also added in the pixel coordinates 5\% of the time to simulate noise in the pipeline. Cross-entropy is used as the loss function with the Adam optimizer. Learning rate is set at a constant $0.01$. For the input normalization, we scale all $x$-pixel coordinates and $y$-pixel coordinates of the feature (court, pose, and shuttlecock) to the interval $[1, 2]$, and set undetected shuttle and pose coordinates to $(0, 0)$. All features were normalized together to ensure that the spatial relationship is preserved. Finally, due to the class-imbalance between hit and no-hit events, we rebalance the dataset to ensure an equal number of each type of event were used in the dataset.

\paragraph{Constrained optimization of the network output}
Next, we incorporate badminton domain knowledge to further optimize the HitNet output. The optimization is based on imposing several constraints. In particular,
\begin{itemize}[noitemsep]
\item[I.] We know the approximate number of hits for a given rally. Based on typical shuttle speeds and the court size, the average time between hits is around 1 second, implying that a rally lasting $D$ seconds has approximately $D$ hits.
\item[II.] No two hits can be too close in time. Empirically, we found half a second to be a good threshold. In the TrackNetV2 dataset, no two hits are within 0.5s of each other.
\item[III.] Hits must be alternating between opposing players, hence no two adjacent hits should be classified to the same player.
\end{itemize}
Our optimization aims to find a set of hits roughly maximizing the sum of confidence scores subject to these constraints.

More formally, given $F$ frames and a set of per-frame confidence scores $\left\{\left(s_1^{(i)}, s_2^{(i)}, s_3^{(i)}\right)\right\}_{i=1}^F$, our goal is to associate each frame with a label $p_i \in \{1, 2, 3\}$ indicating no hit, or a hit by one of the players. Since the three scenarios are mutually exclusive, it suffices to label all the frames on which hits occur (the rest will be labeled as no hits). 

Let $\{t_j\}_{j=1}^M$ denote the frame indices where a hit has occurred, with $M$ total hits. Suppose the total duration of the video is $T$ seconds (implying $T$ hits on average), and the video is at $f$ fps. We seek to maximize the following objective:
\begin{align*}
    \max_{h_{t_j}, t_j} & \sum_{j=1}^M \left( s_{h_{t_j}}^{(t_j)} - \tau \right), \\
    \text{s.t.}\quad & M \leq T, \\
         & t_{j+1} - t_j \geq f/2 \quad \forall 1\leq j<M, \\
         & h_{t_j} \neq h_{t_{j+1}}, \quad h_{t_j} \in \{2,3\} \quad \forall t_j
\end{align*}
where $\tau$ is a parameter that encourages the algorithm to use fewer than $T$ hits if possible. Without $\tau$, the algorithm will always use exactly $T$ hits as none of the confidence scores are negative. In practice, we set $\tau$ to be the mean of $\frac{s_2^{(i)}+s_3^{(i)}}{2}$ across all the frames. The first two inequalities enforces constraint I and II, and the third enforces the shuttle to be hit by alternating players. The global optimum of the objective above can be found by standard dynamic programming running in $O(Tf)$ time.

In \Cref{tab:hit-detection} we compare our model against a naive baseline that simply detects a hit when the second derivative of either the shuttle $x$-coordinate or $y$-coordinate exceeds a predefined threshold (tuned for maximum accuracy). The locations of large second derivative indicate ``discontinuities" in the velocity that occur when a shuttle is struck. To measure the effectiveness of our constrained optimization, we further compare against a naive post-processing which simply classifies using the largest confidence score while ensuring that no two hits are within 0.5 seconds of each other. If two frames are classified as hits and are within 0.5 seconds of each other, the earlier one is classified as a hit and the later one is classified as a no hit.

To measure the accuracy, recall, and precision of the models, we only look at frames where hits occurred. If we were to include all frames, then a trivial detector outputting no hit for all frames would get close to 100\% accuracy. To be precise, suppose the ground truth has hit-player pairs $G=\{(t_i, p_i)\}$ indicating that player $p_i$ hit the shutte on frame $t_i$, and the model predicts $\hat{G} = \{(\hat{t_i}, \hat{p_i})\}$. The metrics we use are:
\begin{equation*}
    \textrm{acc.} = \frac{|G \cap \hat{G}|}{|G \cup \hat{G}|}, \quad \textrm{recall} = \frac{|G \cap \hat{G}|}{|G|}, \quad \textrm{prec.} = \frac{|G \cap \hat{G}|}{|\hat{G}|}.
\end{equation*}
As \Cref{tab:hit-detection} shows, our model offers a substantial ($>35\%$) accuracy improvement of the hit detection over the naive model.

\begin{table}[t]
\caption{{\bf Comparison of HitNet over baselines and ablation study} shows that HitNet benefits from all input features with the optimization-based postprocessing. Derivative-based method attempts to detect hits by thresholding on trajectory derivatives, and RF is a random forest-based classifier. HitNet is our model. \label{tab:hit-detection}}
\centering
\begin{tabular}{l|llll}
\toprule
           & recall & acc. & prec. & f1 \\ 
\hline
Derivative-based     & 65.7\%       & 53.8\%       & 74.7\%        & 0.699 \\ 
\hline
RF                   & 65.1\%       & 57.4\%       & 83.0\%        & 0.730 \\
\hline
HitNet (Shuttle)      & 70.2\%      & 68.8\%       & {\bf 97.4\%}  & 0.815 \\
HitNet (Shuttle+Pose) & 73.9\%      & 75.6\%       & 97.1\%  & 0.850 \\
HitNet (All)          & 78.1\%      & 76.4\%       & 97.2\%  & 0.866 \\
\hline
HitNet+optimization  & {\bf 94.3\%} & {\bf 89.7\%} & 94.9\%        & {\bf 0.946} \\
\bottomrule
\end{tabular}
\end{table}

\subsection{3D Reconstruction}
\label{sec:3d-recon}

With the shots segmented (\S\ref{sec:shot_segmentation}), we can now independently reconstruct trajectories for each shot. We pose this as a constrained nonlinear optimization problem.

\paragraph{Physics-based trajectory estimation} The ability to reconstruct shot-by-shot offers great advantage because without the discontinuous forces applied to the shuttle, the 3D trajectory, $\boldsymbol{x}(t)$, can simply be approximated\footnote{The rotation and spin of the shuttle also affects the motion, which can be accounted for with more complicated models. However, we find the simple drag model to be sufficient for our purposes.} by a particle under drag~\cite{cohen_physics_2015}:
\begin{align}
                       &\diff[2]{\boldsymbol{x}}{t} = \boldsymbol{g} - C_d ||\boldsymbol{x}||_2 \boldsymbol{x}, \\
\text{subject to}\quad & \boldsymbol{x}(0) = \boldsymbol{x}_0, \quad \diff{\boldsymbol{x}}{t}(0) = \boldsymbol{v}_0
\label{eq:drag}
\end{align}
with initial position $\boldsymbol{x}_0 = (x_0, y_0, z_0)^\intercal$, velocity $\boldsymbol{v}_0$, and the drag coefficients $C_d$. $\boldsymbol{g}$ is a constant representing the gravitational acceleration. Given $\boldsymbol{x}_0$, $\boldsymbol{v}_0$, and $C_d$, we can integrate this differential equation to get $\boldsymbol{x}(t)$. Note $C_d$ can change from shot to shot, as the shuttlecock feathers slowly break over the course of a rally. 

\paragraph{Estimating the initial conditions}
The problem for the above equation is of course that the initial conditions are unknown. However, note that given 3D trajectory estimates and camera parameters, we can project $\boldsymbol{x}(t) \in \mathbb{R}^3$ to image space to obtain 2D trajectory estimates $\boldsymbol{\hat{x}}(t) \in \mathbb{R}^2$ using the Direct Linear Transform~\cite{abdel2015direct}. This requires $6$ known 3D coordinates, which we have via the $4$ boundary court corners detected in \S\ref{sec:court-detection} plus the $2$ tip points on the net poles\footnote{The 2D frame position of the pole is found by orthogonally projecting the midpoints of the sidelines up towards the closest white line that is approximately parallel to the back line of the court}. Given the camera parameters, we can measure how good a given 3D trajectory is by measuring the reprojection error:
\begin{align}
\label{eq:reprojection_loss}
    \mathcal{L}_r = \|\boldsymbol{\hat{x}}(t) - \boldsymbol{\tilde{x}}(t)\|^2_2,
\end{align}
where $\tilde{x}$ is the tracked 2D coordinates of the shuttlecock we introduced in \S\ref{sec:shuttle-tracking}. This problem can then be solved with a non-linear regression optimizer until we find a good set of initial conditions.

\paragraph{3D trajectory reconstruction algorithm}

The vanilla version of our reconstruction algorithm is therefore built on solving \Cref{eq:drag}, and refining the initial conditions by reprojecting the solution back to image space using \Cref{eq:reprojection_loss}. 

The reconstruction can be greatly improved by incorporating additional priors. We can provide priors on the start and end positions of the shot through the players' poses. We can also penalize the unlikely event that the shot goes out by extending the trajectory of the shuttle until it hits the ground.\footnote{In professional play, the shuttle rarely goes out by more than a few inches.} The final loss we use is
\begin{align}
    \label{eq:final_3d_loss}
    \mathcal{L}_{3d} = \sigma \mathcal{L}_r + \|\boldsymbol{x}(0) - \boldsymbol{x}_H\|^2_2 + \|\boldsymbol{x}(t_R) -\boldsymbol{x}_R\|^2_2 + d_O^2,
\end{align}
where $\boldsymbol{x}_H$ and $\boldsymbol{x}_R$ are the 3D position estimates of the hitting and receiving players. These are estimated using their feet position at the time of hit and receive, respectively, with a vertical height estimation of \SI{2}{m}. $d_O$ is the distance out of the court if we extend the trajectory of the shuttle until it hits the ground. If the shuttle lands inside the court, $d_O = 0$. We found the use of $d_O$ is quite important, as it helpfully rules out the cases where the shuttlecock shoots out the back or side of the court in a way that cannot be penalized by the reprojection loss. $\sigma$ is used to adjust the closeness of the reprojection to the initial and final coordinate guesses. In practice, we use $\sigma = ||P||_2^{-2}$, where $P$ is the camera projection matrix. This choice of $\sigma$ allows us to compensate for the fact that $\mathcal{L}_r$ is measured in image coordinates, while the other three terms in \Cref{eq:final_3d_loss} are measured in 3D world coordinates. This optimization is solved with some domain-specific constraints:
\begin{itemize}
    \item The initial and final 3D shuttle coordinates should have height less than 3 metres, i.e., $0 \leq x_0 \leq 3$.
    \item The initial velocity of the shuttle is less than 426 kph, or roughly 120 m/s, i.e., $||\boldsymbol{v}_0||_2 \leq 120$.
    \item The initial coordinates of the shuttle is on the same side as the player that hit it. Since the court is 13.4 metres long, this constraint implies that $0 \leq x_0 \leq 6.1$, $0 \leq y_0 \leq 6.7$ if the closer player hit the shuttle, and $6.7 \leq y_0 \leq 13.4$ if the farther player hit the shuttle.
    \item The initial velocity is towards the opposing player, i.e. $\boldsymbol{v}_0^\intercal (\boldsymbol{x}_R - \boldsymbol{x}_H) \geq  0$.
\end{itemize}

\section{Experiments}
\label{sec:experiments}
We have already shown several benchmark comparisons and ablation studies for the court detection, shuttlecock tracking, and shot segmentation (\Cref{tab:hit-detection}) in previous sections. In this section, we focus on experiments around 3D trajectory reconstruction and discuss several factors affecting its accuracy.

\paragraph{Synthetic trajectory dataset}
On top of the dataset introduced in \S\ref{sec:dataset}, we create an additional evaluation dataset containing 10k synthetic trajectories to obtain ground truth 3D positions. We do this by randomly sample the initial conditions and start position, and reject samples that fail to reach over the net or land on the opposite court. With the initial conditions, we can solve \Cref{eq:drag} to obtain the trajectories. Vertical height up to \SI{2.5}{m} and speed up to \SI{1500}{\meter/\second} is used. To ensure even coverage in the simulated trajectories, we simulated trajectories with initial heights up to 2.5 metres, divided the $3.05\times 6.7 \times 2.5$ metre quarter court into $10\times 20 \times 20$ cm cells, and generated a trajectory from each cell. The remaining quarters are symmetric and do not need to be simulated. Our synthetic dataset allows us to measure both the \emph{reconstruction error} (distance between the true 3D trajectory and the reconstructed one) as well as the \emph{reprojection error} (pixel distance between true trajectory and the reconstructed one when projected to 2D).

\begin{figure*}[ht]
    \centering
    \includegraphics[width=0.9\textwidth]{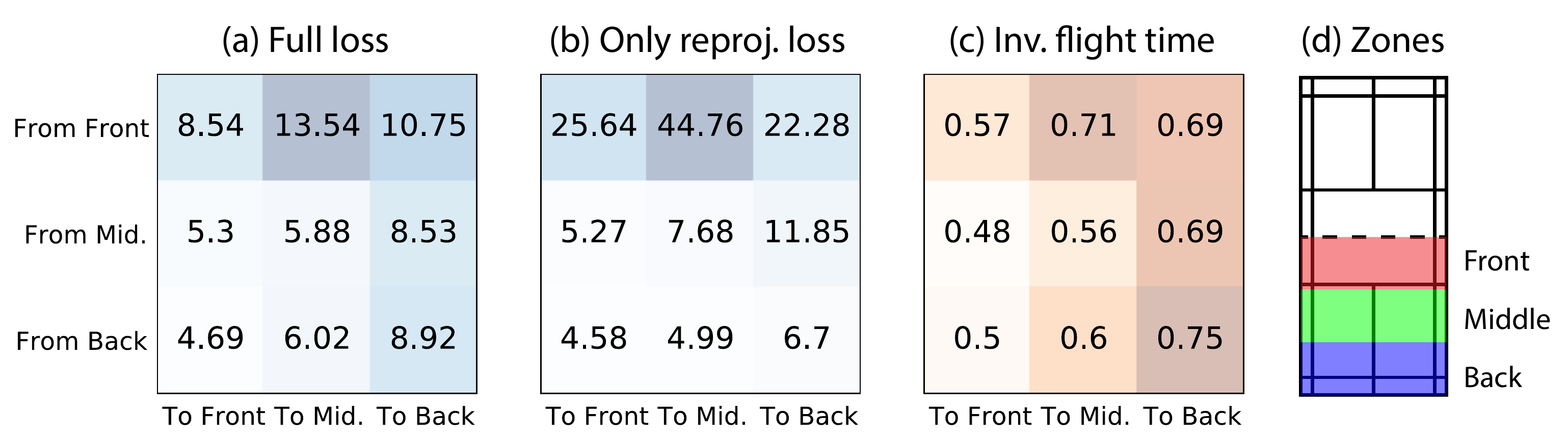}
    \caption{{\bf Measured reconstruction error (in cm) on 10k synthetic trajectories} shows that shots leaving from the front cause larger errors. (a) is based on optimizing our full loss \Cref{eq:final_3d_loss}, where (b) is based on optimizing only the reprojection loss \Cref{eq:reprojection_loss}. (c) shows the inverse flight time of the shuttle for each zone, and (d) shows the standard badminton court and the zone assignment, where we have divided the half-court into front, middle, and back zones covering one-third each.
    \label{fig:simulated-reconstruction}}
\end{figure*}

\paragraph{Reconstruction accuracy on synthetic data}

We first evaluate our reconstruction algorithm on the simulated trajectories. Using camera projection matrices from the three test matches, we project the 3D coordinates onto the image space and reconstruct it back using our algorithm. To mimic uncertainty that might occur in the pipeline affecting the estimated impact location, we add uniformly distributed random noise between \SI{-0.5}{m} and \SI{0.5}{m} onto the hit positions, $\boldsymbol{x}_H$ and $\boldsymbol{x}_R$. In \Cref{fig:simulated-reconstruction}, we show the accuracy of our reconstruction on this synthetic dataset. We bundle trajectories that start and end in different zones to illustrate the effect camera perspective and travel distance might have on the error (further discussion on this in later sections). On average, adding the priors improves the error from \SI{14.9}{cm} to \SI{8.0}{cm}. The 2D reprojection error are not shown as they are consistently less than $1$~pixel for all zones, showing our algorithm is working well to minimize the reprojection error.

\paragraph{Reconstruction accuracy on real data}

We use the dataset introduced in \S\ref{sec:dataset} that contains real-world matches to study the reconstruction accuracy. Note that in this dataset, 3D ground truth positions are not available and thus only 2D reprojection error can be computed. To tease out effects of different parts of the pipeline, we perform two versions of the study: one where we use ground truth shuttle tracking and hits, and the other where all features were estimated with the system. The result is shown in~\Cref{tab:real-reconstruction}. As expected, the estimated shuttle tracking and hits do contain errors, and thus the end-to-end reconstruction contains up to four times the error in pixel counts. The highest error is $37.1$~pixels, or about $5\%$ given the image resolution we are working with. For the measurement, we exclude the first and last shot of a rally to account for the often occluded first shot (serve), and the last shot (ground impact) as they are not annotated.

\begin{figure}[ht]
    \centering
    \includegraphics[width=0.8\columnwidth]{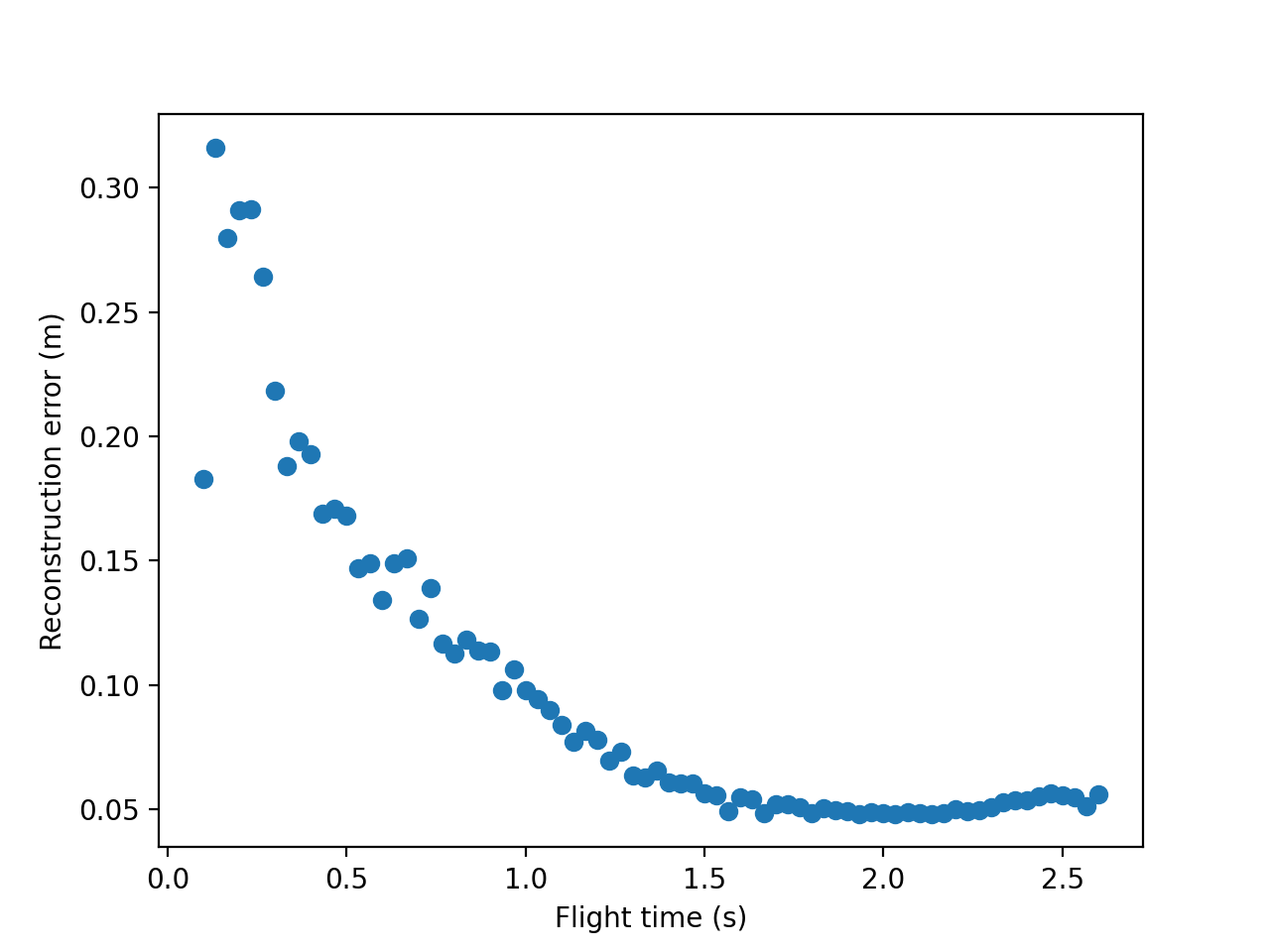}
    \caption{{\bf Reconstruction error with respect to shuttlecock flight time} shows that longer flight time leads to drastically lower error until saturated at around \SI{5}{cm}.
    \label{fig:recon_vs_flight}}
\end{figure}

\paragraph{Error attribution}
Inspection of the performance on both synthetic and real data reveals several observations regarding the reconstruction error: 1) the error tends to be higher when a shot is leaving from the front court, and demonstrates certain zone specific behaviors; 2) misclassification of a hit (either false positive or false negative) has a rippling impact on the reconstruction; 3) certain limitations of the aerodynamic model in simulating the true trajectory.

As shown in \Cref{fig:simulated-reconstruction}, the reconstruction errors are always higher when the shots are leaving from the front court. In \Cref{fig:recon_vs_flight}, we show that the error is highly correlated with the shuttlecock flight time. This is because longer trajectories naturally correspond to more on-screen observations, which in turn constrain the optimizer to find a more accurate initial conditions. Similarly, the inverse flight time shown in~\Cref{fig:simulated-reconstruction} shows good agreement with this observation. We also note that when shuttlecock flies so high that it becomes off-screen, typically in a to-the-back shot, the number of observations are reduced, too. This helps explain why shots going to the back court also results in higher error.

We have briefly discussed the observation in \Cref{tab:real-reconstruction} that the end-to-end reconstruction contains much higher errors than the reconstruction bootstraped by the ground truth shuttle and hit labels. First, for the error in the bootstraped reconstruction, visual inspection of the pose and the court detection results show that they are sufficiently accurate, so we believe this might be due to the approximations made in the simplified aerodynamic drag model we used in \Cref{eq:drag}. Shuttlecock can experience changing cross sectional area and thus changing drag coefficients \emph{during} a shot; it can tumble and flip dynamically when a front-to-front net shot is played; some players ``slice" the shuttlecock harder, making the spin of the shuttlecock another variable that is not modeled. An improved aerodynamic model can improve the baseline performance.

On the other hand, the error in the end-to-end reconstruction is largely due to misclassified hits. Although both false positive and false negative will disrupt the model, failure to detect or misclassifying the player who hit the shot will have severe consequences. In the former case, not detecting a shot means that the dynamic model in \Cref{eq:drag} is invalid as the near-instantaneous energy input from the racket is suddenly present. It is therefore not surprising that reconstruction will fail. In the latter case, misclassifying the player who hits the shot might cause our postprocessing algorithm to completely misplace the sequence, causing the entire rally to be ruined. 

Combining these two effects, even though our shot segmentation has around $90\%$ accuracy, this can result in erroneous reconstruction for $20\%$ of the shots, as a hit is sandwiched by two shots. Therefore, it is of paramount importance that the hit detection model be improved; we leave this to future work.

\begin{table}
\caption{{\bf Measured reprojection error on the real data} for both bootstrapped pipeline using ground truth shuttlecock tracking and hit detection, and the end-to-end pipeline where all features were inferred. This table reveals that even with bootstrapped labels the reconstruction is still not perfect. On the other hand, incorporating inferred shuttlecock tracking and hits will significantly increase the error.\label{tab:real-reconstruction}}
\centering
\begin{tabular}{l|l|l}
\toprule
           & Error (bootstrapped) & Error (end-to-end) \\ 
\hline
Match 1 & 8.1 px & 37.1 px \\
Match 2 & 8.9 px & 28.8 px \\
Match 3 & 9.8 px & 23.3 px \\
\bottomrule
\end{tabular}
\end{table}

\section{Conclusion}

In this paper, we introduce a novel shot segmentation and 3D trajectory reconstruction method for monocular badminton videos. To segment the shots, we leverage domain-specific court, pose, and tracked shuttlecock positions to design an efficient GRU-based recurrent network that achieves $90\%$ accuracy on an enhanced TrackNet dataset. Using these shots, we show that it is possible to pose the monocular reconstruction problem as nonlinear optimization with the help of a physics-based dynamic model. Finally, we evaluate our method on both synthetic data and real data, and discuss its strength and weakness in relation to shuttlecock flight time, as well as the starting and ending position of shots.

Our method has several avenues for future exploration. To increase the robustness of the system, we are currently in the process of annotating additional data. We believe more data, especially those of different views, can greatly improve the robustness of the system. Another extension of our hit detection is shot type classification. Given shot type annotations along with the hits, we believe it is possible to build a robust shot-type in a manner similar to our hit detector. We note that our reconstructed 3D trajectories can have many downstream applications such as shot retrieval, novel view synthesis, highlight detection, and statistics gathering. Finally, we note that although we are demonstrating our method's efficacy on badminton videos, we expect it to generalize to other racket sports. 

\pagebreak


{\small
\bibliographystyle{ieee_fullname}
\bibliography{references}
}

\end{document}